\documentclass[lettersize,journal]{IEEEtran}
\usepackage{amsmath,amsfonts}
\usepackage[ruled,vlined]{algorithm2e}
\usepackage{array}
\usepackage[caption=false,font=normalsize]{subfig}
\usepackage{textcomp}
\usepackage{stfloats}
\usepackage{url}
\usepackage{verbatim}
\usepackage{graphicx}
\usepackage{cite}
\usepackage{tcolorbox}
\usepackage{framed}
\begin{document}

\newcommand{\xl}[1]{{\color{orange}{[xl:#1]}}}
\newcommand{\fei}[1]{{\color{purple}{[fei:#1]}}}

\title{Large Language Model Aided Multi-objective Evolutionary Algorithm: a Low-cost Adaptive Approach}


\author{Wanyi Liu,~Long Chen,~Zhenzhou Tang,~\IEEEmembership{Senior~Member,~IEEE}

\thanks{Y. Liu and Z. Tang are with the Wenzhou Key Laboratory for Intelligent Networking, Wenzhou University, Wenzhou, China, 325035.}
\thanks{L. Chen is with the Key Laboratory of Intelligent Education Technology and Application of Zhejiang Province, Zhejiang Normal University, Jinhua, China, 321000.}
\thanks{This work was supported by the Natural Science Foundation of Zhejiang Province, China, under Grant LZ20F010008.}}

\markboth{Journal of \LaTeX\ Class Files}%
{Shell \MakeLowercase{\textit{et al.}}: A Sample Article Using IEEEtran.cls for IEEE Journals}


\maketitle

\begin{abstract}

Multi-objective optimization is a common problem in practical applications, and multi-objective evolutionary algorithm (MOEA) is considered as one of the effective methods to solve these problems. However, their randomness sometimes prevents algorithms from rapidly converging to global optimization, and the design of their genetic operators often requires complicated manual tuning. To overcome this challenge, this study proposes a new framework that combines a large language model (LLM) with traditional evolutionary algorithms to enhance the algorithm's search capability and generalization performance.In our framework, we employ adaptive and hybrid mechanisms to integrate the LLM with the MOEA, thereby accelerating algorithmic convergence. Specifically, we leverage an auxiliary evaluation function and automated prompt construction within the adaptive mechanism to flexibly adjust the utilization of the LLM, generating high-quality solutions that are further refined and optimized through genetic operators.Concurrently, the hybrid mechanism aims to minimize interaction costs with the LLM as much as possible. By adopting this approach, we maximize the benefits of the LLM’s language understanding and generation capabilities, providing the MOEA with more intelligent and efficient search strategies. We conducted experiments on multiple multi-objective optimization benchmark problems, and the results demonstrate that LLM-assisted evolutionary search significantly accelerates population convergence, allowing it to stand out in competition with state-of-the-art evolutionary algorithms. These experimental findings reveal the potential advantages of utilizing pre-trained LLMs in the design of MOEAs. This framework combines LLM with MOEA, offering a promising new approach to solving multi-objective optimization problems. We believe that as LLM continue to evolve and improve, they will play an increasingly significant role in multi-objective optimization and other complex decision-making challenges.
\end{abstract}


\section{Introduction}
\IEEEPARstart {T}{he} multi-objective optimization problem refers to the optimization problem which needs to consider several interrelated and mutually restricted optimization objectives in the decision-making process. Multi-objective optimization problems are very common in the real world. The key challenge to solve the multi-objective optimization problem is how to optimize multiple interdependent objectives to efficiently find the feasible Pareto optimal solution.

Multi-objective evolutionary algorithms (MOEAs), which draw inspiration from the evolutionary processes observed in nature, have emerged as essential methodologies for addressing multi-objective problems. Nonetheless, conventional MOEAs necessitate the formulation of sophisticated operators to enhance their efficacy, including the refinement of solution convergence and distribution. This endeavor represents a significant and labor-intensive challenge. Enhancing the convergence rate, solution diversity, and reducing the dependency on expert knowledge in MOEAs are pivotal to augmenting their applicability in real-world optimization scenarios.

Recent years have witnessed increasing attention on the integration of Large Language Models (LLMs) with the MOEA framework, with the aim of enhancing the performance and adaptability of MOEA algorithms through the advanced natural language understanding and generation capabilities inherent in LLMs. LLMs present two principal application patterns for optimization problem-solving. The first involves leveraging LLMs as black-box search operators, employing their representation and generation capacities to efficiently navigate the solution space and iteratively generate successive solution iterations, without the need for explicit problem encoding ~\cite{introductionLLM1_1,introductionLLM1_2,introductionLLM1_3}. The potential of LLMs in this domain has been corroborated by numerous studies, showcasing their robust optimization capabilities in addressing small-scale optimization problems. Unlike traditional MOEA methodologies, this approach allows the use of natural language to articulate the desired characteristics of an optimization problem's solution, thereby simplifying the process compared to programming or mathematical descriptions. This method also minimizes the necessity for expert intervention, facilitating broader generalization across diverse problems.

Another approach leverages the extensive prior knowledge of LLM to design high-performance operators or select appropriate operators for algorithm evolution, thereby enhancing the search capability and convergence speed of the algorithm. LLM possesses the capability to select or generate the most suitable algorithm for a specific optimization problem based on its profound understanding, interpret the natural language description of the problem, and subsequently formulate optimization strategies ~\cite{introductionLLM1_4,introductionLLM1_5}. In this context, LLM's code comprehension, representation, and generation capabilities are primarily utilized to augment the performance of the MOEA algorithm. Notably, LLM is not employed directly as a search operator within the optimization process but represents a significant advancement at the algorithmic level.

Nevertheless, how to combine LLMs with the MOEA framework in a way that makes them applicable to most MOEA frameworks remains an open issue. To address this, this paper proposes a novel low-cost adaptive framework where LLMs assist MOEAs.

Specifically, this framework utilizes an auxiliary evaluation function to assess the current population state, allowing for selective use of the LLM to accelerate the algorithm's convergence. When generating solutions with the LLM, a small number of elite solutions generated by the traditional MOEA algorithm are used as examples to construct prompts, enabling the LLM to generate superior Pareto-optimal solutions. 
Then, by mixing the LLM-generated Pareto solutions with the elite solutions from traditional MOEA, the framework applies traditional MOEA algorithms for crossover and mutation, thereby accelerating the convergence speed and increasing the diversity of the population.

Compared to existing methods, the advantage of this framework is twofold.
On one hand, it incorporates a flexible adjustment mechanism. 
The framework employs an adaptive mechanism to determine when to use the LLM based on the auxiliary evaluation function designed in this paper, ensuring that LLM is enabled only when LLM has a significant contribution to the search process, thereby minimizing interaction costs. At the same time, through the pre-set prompt construction algorithm, the prompt samples can be selected flexibly and the prompt of each iteration can be generated systematically without manual participation, and it can ensure that the model can be correctly guided to generate a solution closer to the Pareto preface. On the other hand, a hybrid mechanism is adopted. Large language models only need to generate a small number of solutions, minimizing the number of interactions with LLMS. Compared with methods that rely on LLMS for each iteration, this framework minimizes computation time and cost as much as possible. It can not only fully exert the performance of MOEA itself, but also use LLM's own language understanding and generation ability to improve the generalization ability, convergence speed and generation ability of the algorithm. At the same time, the hybrid mechanism and the combination mechanism reduce the threshold of the combination of MOEA framework and LLM, and improve the applicability of the framework.

Our main contributions are summarized as follows:

\begin{itemize}

    \item Low cost search strategy. The framework uses a hybrid strategy that leverages LLM to generate a small number of high-quality solutions, minimizing the computational and time costs associated with LLM interactions. The algorithm can make full use of the reasoning and searching capabilities of LLM without incurring too much overhead.

    \item Flexible adjustment mechanism. Our framework uses auxiliary evaluation functions to tailor the use of LLM to ensure that its assistance to large language models is maximized. At the same time, pre-defined templates are used to iteratively generate hints, ensuring that the LLM receives clear and indirect instructions and avoids manual labor. In addition, auxiliary evaluation functions and prompt templates can be easily customized for different MOEA algorithms and optimization problems, enhancing the adaptability and versatility of the framework.

    \item general applicability. The modular design of our framework allows for the easy integration of various MOEA algorithms, and this adaptability ensures that our framework can be applied to a wide range of multi-objective optimization problems, providing a valuable tool for researchers and practitioners.
    
    \item Experiments on multiple multi-objective optimization benchmark problems show that LLM-assisted MOEA can achieve better performance than the widely used MOEA in solving multi-objective optimization problems.
    
\end{itemize}

\section{Related Works}

\subsection{MOEA using the LLM}
Over the past two years, we have witnessed the rapid development of LLMS, , with their scale growing exponentially. Thanks to massive training datasets, these models have become increasingly powerful.  LLMs have demonstrated exceptional performance in fields such as natural language processing, data analysis, and other areas of scientific research~\cite{relatedWorkMOEALLM1}, and have made significant progress in providing personalized services~\cite{relatedWorkMOEALLM2}.  By leveraging vast amounts of data, LLMs have also shown remarkable capabilities in reasoning and prediction. The flexibility of these models allows them to tailor solutions for a variety of optimization problems, exhibiting superior generalization abilities. Against this backdrop, researchers have begun to explore the potential of combining LLMs with evolutionary computation techniques.

Current research advancements mainly include two approaches: using LLMs as black-box operators for optimization problems~\cite{relatedWorkMOEALLM3}, and selecting or generating optimization algorithms suitable for specific problems based on the strong understanding and generation capabilities of LLMs~\cite{relatedWorkMOEALLM9,relatedWorkMOEALLM10,relatedWorkMOEALLM11,relatedWorkMOEALLM12}.

The use of LLMs as black-box search operators for optimization problems is a current research hotspot, with progress summarized in Table~\ref{table
}. Due to the interactive nature of LLMs, the initial integration focused on single-objective problems. Yang et al. proposed the Optimizer Prompting (OPRO) method, which leverages LLMs as optimizers to solve single-objective optimization problems~\cite{relatedWorkLLM3}. This algorithm is particularly suited for cases where gradients are unavailable. Natural language is used to describe the optimization problem and serve as meta-prompts. The primary role of the LLM is to use previously generated solutions as prompts to continually generate new solutions, which are then evaluated and added to the prompts for the next iteration.
Liu et al. proposed the LLM-driven Evolutionary Algorithm (LMEA)\cite{relatedWorkLLM4}, where LLMs are employed to perform crossover and mutation operations. Prompts are constructed in each generation to assist the LLM in selecting parent solutions. As research has progressed, attention has shifted towards multi-objective optimization problems. Liu et al. used a decomposition-based MOEA framework as a backbone\cite{relatedWorkMOEALLM6}, with the LLM acting as a black-box optimization operator. Through prompt engineering and contextual learning, the LLM generates new offspring solutions for each subproblem. To mitigate the high interaction costs of using LLMs, they further designed an explicit white-box linear operator to approximate the results of the LLM.

LLMs can also be applied to quality-diversity (QD) search~\cite{relatedWorkMOEALLM7}. Bradley et al. proposed the QD-based AI Feedback (QDAIF) algorithm to address this~\cite{relatedWorkMOEALLM8}, which can be applied to complex qualitative domains. This algorithm uses LLMs to evaluate both the quality and diversity of solutions, while an evolutionary algorithm is responsible for maintaining a solution archive. Higher-quality and more diverse solutions, as assessed by the LLM, are then introduced into the solution set.
Although these studies have introduced new possibilities, the application of LLMs to the design of optimization algorithms remains a relatively new field, still in its early exploratory stage. So far, these studies have primarily applied LLMs as black-box solvers for optimization problems, which means that each step of the optimization process requires real-time, resource-intensive online interaction with the LLM. This mode of interaction can result in significant consumption of computational resources and time. Moreover, applying LLMs to complex real-world optimization problems presents numerous challenges. The scope of current evaluation studies is relatively narrow, considering only a limited set of factors, and thus has yet to comprehensively validate the overall capabilities of LLMs in the optimization domain. Overall, despite encouraging preliminary findings, there remain significant challenges to effectively applying LLMs to complex optimization problems in real-world scenarios.

\begin{table*}[htbp]
\centering
\caption{The LLM assists in the summary of MOEA's work}\label{table:relatedWork}
\resizebox{0.85\textwidth}{!}{%
\begin{tabular}{ccccccc}
\hline
Index & References & Type & Methodology      \\
\hline
1     & ~\cite{relatedWorkLLM2} & N/A,Evaluation & AS-LLM \\
2     & ~\cite{relatedWorkLLM3} & Sing-objective & OPRO \\
3     & ~\cite{relatedWorkLLM4} & Sing-objective & LMEA \\
4     & ~\cite{relatedWorkMOEALLM6} & Multi-objective,MOEA/D & MOEA/D-DE,MOEA/D-LO \\
5     & ~\cite{relatedWorkMOEALLM8} & Multi-objective,Evaluation & Decomposition-based MOEA QDAIF \\
\hline
\end{tabular}%
}
\end{table*}

\section{Problem Formulation}

For the sake of generality, this paper considers the following form of multi-objective minimization problem:
\begin{equation}
    \begin{aligned}
    \text{min. } & F(\mathbf{x})=(f_{1}(\mathbf{x}), \ldots, f_{m}(\mathbf{x}))^T,\\
    \text { s.t. } & \mathbf{x} \in \Omega,
    \end{aligned}
\end{equation}
where $\mathbf{x}=\{x_1,\dots,x_n\}^T$ is the decision variable, and $\Omega$ is the domain of the decision variable, usually a $n$ dimensional Euclidean space. $F(\mathbf{x})$ is an objective function vector that contains $m$ of objective functions, which usually represent different performance metrics or costs that need to be considered simultaneously during optimization. The goal of a multi-objective optimization problem is to find a set of solutions that can not be improved simultaneously on all objective functions, that is, the Pareto optimal solution set. These solutions represent an optimal trade-off between different goals and can provide decision-makers with multiple options to make decisions based on specific needs and preferences.

In multi-objective optimization problems, a solution ${x}_i$ is said to dominate another solution ${x}_j$ if the following two conditions are simultaneously satisfied: 
\begin{equation}
    \begin{aligned}
     &  f_{m}(x_i) \leq f_{m}(x_j), &\forall m \in \{1,2 \ldots M\},\\
     &  f_{n}(x_i) < f_{n}(x_i), &\exists n \in \{1,2 \ldots M\}. 
    \end{aligned}
\end{equation}
In this case, we say that individual $x_i$ dominates individual $x_j$, denoted as $x_i \prec x_j$.
If a solution $\mathbf{x}^*$ is not dominated by any other solution $\mathbf{x} \in \Omega$ (the feasible solution set), then $\mathbf{x}^*$ is called a Pareto-optimal solution~\cite{miettinen2012nonlinear}.
The set of all Pareto-optimal solutions is referred to as the Pareto set. Correspondingly, the Pareto front (PF) is the set of objective function values corresponding to all Pareto-optimal solutions. In the graphical representation of multi-objective optimization, the Pareto front is typically a curve or polyhedron that illustrates the trade-off relationships between different objectives. In multi-objective optimization, the goal is to approximate the entire Pareto frontier as uniformly as possible.

\section{LLM for MOEA}

\subsection{Framework}
\label{framework}
\begin{figure*}[!t]
    \centering
    \includegraphics[width=0.9\linewidth]{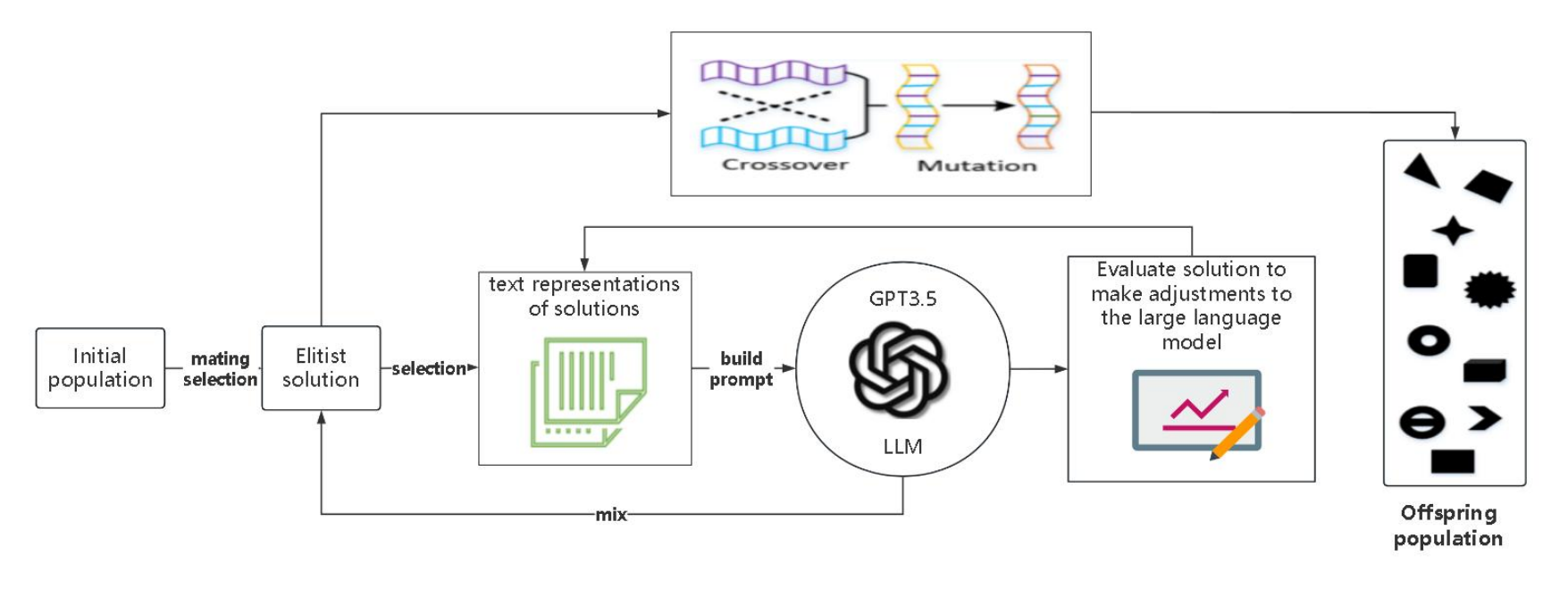}
    \caption{The LLM-assisted MOEA framework integrates LLMs as a black-box optimization operator. The LLM generates high-quality solution sets through prompts, accelerating the algorithm's convergence while also increasing the solution diversity. Prompt engineering is used to facilitate contextual learning for LLM.}
   \label{fig:framework}
\end{figure*}

In this section, we will present the LLM-assisted evolutionary framework proposed in this paper. This framework is a general-purpose framework that can be integrated with any other multi-objective evolutionary algorithm to enhance the convergence and distribution of the algorithm. In the following, we will take NSGA-II~\cite{Nsgaii} as an example and explain the overall design of the framework by combining the NSGA-II algorithm process and operators, as illustrated in Figure~\ref{fig:framework}.

During the initialization stage, an initial population $\mathcal{P}_t = \{x_1, x_2, \ldots, x_N\}$ is randomly generated, where $N$ denotes the population size and $x_k$ represents the $k$th individual in the population. Each individual $x_k = \{x_{k1}, x_{k2}, \ldots, x_{kd}\}$ is a $d$-dimensional decision vector. To ensure that each individual is uniformly distributed within the domain of the decision variables, the NSGA-II algorithm adopts a random generation method for population initialization, namely: 
\begin{equation}
    x_{ki}=l_i+(u_i-l_i)\times r_{ki},\forall i=\{1,2\ldots d\},
\end{equation}
where $x_{ki}$ represents the $i$-th decision variable of individual $x_k$, $l_i$ and $u_i$ are the lower and upper bounds of $x_{ki}$, respectively, and $r_{ki}$ is a random number that follows a uniform distribution in the interval $(0,1)$.

Each individual represents a potential solution to this multi-objective optimization problem. We evaluate each individual in the population, which is to say
:
\begin{equation}
    F(\textbf{x}_k)=\{f_1(\textbf{x}_k), f_2(\textbf{x}_k)\ldots f_M(\textbf{x}_k)\},
\end{equation}
where $M$ denotes the number of optimization objectives, and $f_m(x)$ is the evaluation function. Subsequently, we perform non-dominated sorting on the individuals within the population ~\cite{deb1995multi}, partitioning the population $\mathcal{P}$ into multiple layers $\textbf{F} = \{\textbf{F}_1, \textbf{F}_2, \ldots, \textbf{F}_c\}$, based on the dominance relationships between individuals as follows:
\begin{equation}
    \begin{aligned}
     &  \textbf F_1=\{x_i \in \mathcal{P} | \forall x_j \in \mathcal{P},x_j \not\prec x_i\},\\
     &  \textbf F_{n} = \{x_i \in \mathcal{P}_{n-1} | \forall x_j \in \mathcal{P}_{n-1},x_j \not\prec x_i\},\\
     &  \mathcal{P}_{n-1}=\mathcal{P}\setminus \cup_{k=1}^{n-1} \textbf F_k,n=\{2,3 \ldots c\}.
    \end{aligned}
\end{equation}

Next, we proceed with the selection of evolutionary operators, which involves determining whether to employ LLM to generate new individuals. This decision is made through the auxiliary evaluation function. This section takes NSGA-II as the basic algorithm, and thus the auxiliary function (which can be freely configured) is set as follows:
\begin{equation}
    \begin{aligned}
     &  \textbf D_v=\{d_i \in \textbf{D} | d_i \text{ is finite} \},\\
    & \bar{d} = 
    \begin{cases} 
      \infty, & \text{if } \textbf D_v = \emptyset \\ 
      \frac{1}{|\textbf D_{v}|} \sum_{d_i \in \textbf D_v} d_i, & \text{otherwise} 
    \end{cases},\\
     &  S=-\bar{d}+\frac {1}{|\textbf{B}|} \sum_{f_i \in \textbf{B}} f_i\\
    \end{aligned}
\end{equation}
where $\textbf{D}_v$ represents a subset of the set $\textbf{D}$, which consists of all the finite elements within the set $\textbf{D}$, with $\textbf{D} = \{D_1, D_2, \ldots, D_N\}$. $D_i$ denotes the crowding distance of individual $x_i$, which is calculated as follows:
\begin{equation}
    \begin{aligned}
     &D_i = \sum_{m=1}^{M} \frac{f_m(x_{i+1}) - f_m(x_{i-1})}{f_m^{\max} - f_m^{\min}},\\
    \end{aligned}
\end{equation}
where $ f_m(x_{i+1}) $ and $ f_m(x_{i-1}) $ represent the objective function values of the two neighboring individuals of $ x_i $ with respect to the $ m $-th objective, and $ f_m^{\max} $ and $ f_m^{\min} $ are the maximum and minimum values of the $ m $-th objective function, respectively. The set $ \textbf{B} = \{B_1, B_2, \ldots, B_N\} $, where $ B_i $ is the index of each individual within its corresponding rank $ F_i $.

\begin{algorithm}[ht]
    \caption{Algorithm Framework}
    \label{alg:framework}
    \KwIn{The maximum number of evaluations: $N_{\text{max}}$; The number of population: $N$; The number of parents $l$ for LLM; The number of new individuals $s$ generated by LLM;Decision threshold: $\delta$.}
    \KwOut{Next generation population $\mathcal{P}_{t+1}$.}
    
    
    Generate the initial population $\mathcal{P}_t \leftarrow \{\mathbf{x}_1, \dots, \mathbf{x}_N\}$ randomly;
    
    $\mathcal{P}_{t+1}\leftarrow \emptyset$,$S_{t-1} \leftarrow 0$,$t \leftarrow 0$;
    
    \While{$t<N_{max}$}{
         \textbf{assession:} get score $S_{t}$ by auxiliary function;

        \If{$S_t - S_{t-1} > \delta$}{
            \textbf{Selection:} get elitist solutions $\mathbf{p}_t \leftarrow \{\mathbf{x}_1, \dots, \mathbf{x}_l\}$ through MOEA's own selection algorithm;
        
            \textbf{Prompt engineering:} generate textual $Prompt$ for LLM given the subset $\mathbf{p}_t$ by the method provided in Sec.~\ref{llm_for_optimization};
            
            \textbf{Direction:} LLM generates a set of new individuals $\mathbf{o}_t \leftarrow \{\mathbf{x}_1, \dots, \mathbf{x}_s\}$ with the $Prompt$;
            
            \textbf{Reproduction:} replace the solution in $\mathbf{p}_t$ with $\mathbf{o}_t$, and use  MOEA's own propagation algorithm to generate offsprings $\mathcal{Q}_t \leftarrow \{\mathbf{x}_1, \dots, \mathbf{x}_N\}$;
        }
        \Else{
            \textbf{evolution:} use  MOEA's own propagation algorithm in $\mathcal{P}_t$ to generate offsprings $\mathcal{Q}_t \leftarrow \{\mathbf{x}_1, \dots, \mathbf{x}_N\}$;
        }
        \textbf{Update:} merged population $\mathcal{U}_t \leftarrow \{\mathcal P_t \cup \mathcal Q_t\}$,next generation population $\mathcal{P}_{t+1} \leftarrow  \{\mathbf{x}_1, \dots, \mathbf{x}_N\}$ were selected in $\mathcal{U}_t$ by MOEA child selection algorithm;

        $S_{t-1} \leftarrow S_{t}$, $\mathcal{P}_{t}\leftarrow \mathcal{P}_{t+1}$,$t \leftarrow t + 1$;
    }
\end{algorithm}

We employ a comparison mechanism based on the auxiliary function's score to decide whether to utilize a large language model during the generation of the next population. Specifically, we compare the score $ S_{t-1} $ obtained from the previous generation's population $ \mathcal{P}_{t-1} $ with the score $ S_t $ of the current population $ \mathcal{P}_t $. The method for generating offspring is determined as follows: 
\begin{equation}
    \begin{cases} 
      \text{NSGA-II}, & \text{if } S_t - S_{t-1} < \delta \\ 
      \text{LLM}, & \text{if } S_t - S_{t-1} \geq \delta
    \end{cases},
\end{equation}
where $\delta$ is a pre-defined decision threshold.

\textbf{Generating Offspring Using LLM:} During the process of generating offspring using the LLM, we first employ the tournament selection strategy to select $ N $ individuals from the current population, forming the set $ \mathcal{M}_t $. Within $ \mathcal{M}_t $, we further identify individuals with a high frequency of occurrence, which to some extent represent the characteristics of elite solutions. We rank the individuals based on the frequency of occurrence from high to low and select the top $ l $ individuals to form the set $ \mathbf{p}_t $. We consider that the individuals in $ \mathbf{p}_t $ contain important genetic information, and therefore, we use this information to construct prompts that are input into the large language model. This allows the model to learn and simulate the characteristics of these high-quality individuals.We then replace the recurring individuals in $ \mathcal{M}_t $ with $ \mathbf{o}_t $ and use the existing breeding strategies of the NSGA-II algorithm to generate the offspring $ \mathcal{Q}_t $. Finally, we combine the original population $ \mathcal{P}_t $ with the offspring $ \mathcal{Q}_t $ and use the environmental selection mechanism of the NSGA-II algorithm to get the new generation population $ \mathcal{P}_{t+1} $.

If the LLM is not employed, the offspring $ \mathcal{Q}_t = \{\mathbf{x}_1, \dots, \mathbf{x}_N\} $ are directly generated through the reproductive strategy implemented in the NSGA-II algorithm. The remaining steps are the same as above.

\begin{table}[t]
\centering
\caption{The number of tokens consumed by a single algorithm}\label{table:LLMtoken}
\resizebox{0.9\linewidth}{!}{%
\begin{tabular}{ccccccc}
\hline
Index & Type & LLM Token Count      \\
\hline
1     & Adaptive use of LLM & 56620 \\
2     & LLM is used for each iteration & 283100 \\
\hline
\end{tabular}%
}
\end{table}
The framework integrates the advantages of LLM and EA through adaptive adjustment mechanism and hybrid mechanism. As shown in Table~\ref{table:LLMtoken}, the adaptive adjustment mechanism can not only make full use of the LLM with rich knowledge to help EA effectively search the solution space, but also reduce the cost of interacting with the LLM as much as possible to maximize the benefit. The hybrid mechanism leverages the inherent characteristics of EAs to provide knowledge for LLM  to guide the search process and further improve the speed and quality of convergence. These features give the framework general applicability, allowing users to easily integrate various algorithms and auxiliary evaluation functions, thus making it applicable to a wide range of optimization problems. Consequently, this framework holds significant importance in practical applications, as it can help solve complex optimization problems in the real world, enhancing the accuracy and efficiency of decision-making.

\subsection{LLM as a black-box optimizer}
\label{llm_for_optimization}
In this section, we will focus on describing how to utilize Large Language Models (LLMs) as black-box optimizers, including the selection of prompt samples, prompt construction and the handling of exceptions.

In the process of overcoming token length and runtime constraints, we have found that guiding LLM to generate high-quality solutions using a small number of instances can effectively reduce the interaction costs with LLM. However, due to the limited number of instances and the need for the generated solutions to be as excellent as possible, we need precise prompts to guide LLM. Next, we describe in detail the specific steps for prompt generation.

\subsubsection{Prompts the desired individual for selection}
In the choice of instances, we need to provide high-quality instances to help LLM learn the characteristics of them.The number of selected individuals can be flexibly adjusted due to the interactive flexibility of the LLM.

\subsubsection{Prompt engineering}

We have divided the prompt into 4 parts to make it easy for the LLM to understand and respond quickly:
\begin{itemize}
    \item \textbf{Identity localization:} Due to the broad scope of knowledge encompassed by large language models, clearly defining the role in the domain of multi-objective optimization can enhance its logical reasoning efficiency and rigor.
    \item \textbf{Task description:} A brief description of the task, and a brief overview of the nature of the input information.
    \item \textbf{Context information:} The format of the input information and sample information..
    \item \textbf{Expected output:} Output requirements and format.
\end{itemize}

Examples of prompt are presented below, including: 1. Identity localization: This part positions LLM as an expert in the field of multi-objective optimization, ensuring its accurate comprehension of the subsequent content. 2. Task description: This part provides a concise overview of a minimization problem involving multiple variables, along with relevant definitions and contextual examples. 3. Context information: Describe the format and information of the sample, such as the number of parameters per sample, starting with <start> and ending with <end>. 4, Expected output: specify the format of the LLM output, each output sample begins with <start>, <end> end. This formatted prompt can improve the understanding of the LLM and to facilitate the user to extract the results.

\begin{framed}
\noindent \textbf{Example Prompt:}\\
You are an expert in multi-objective optimization algorithms. Your task is to generate improved solutions with better objective values through given solutions. 

I have several solutions, all of which are in the form of 10 dimensional decision vectors.The following is the initial solution in the mating pool.   \\
solution: \textless{}start\textgreater{}0.322,0.947,0.378,0.583\textless{}end\textgreater{} \\
 obj\_value: 5.483 \\
solution: \textless{}start\textgreater{}0.937,0.264,0.472,0.473\textless{}end\textgreater{} \\
 obj\_value: 3.483 \\
... \\
solution: \textless{}start\textgreater{}0.573,0.483,0.937,0.937\textless{}end\textgreater{} \\
 obj\_value: 1.374 \\
solution: \textless{}start\textgreater{}0.837,0.374,0.374,0.196\textless{}end\textgreater{}   \\
 obj\_value: 1.037   \\
 You can use these multi-objective optimization algorithms to generate new solutions (one or more algorithms can be used). Simply output three new solutions with better objective values. Each solution must start with \textless{}start\textgreater{} and end with \textless{}end\textgreater{}. 
\end{framed}

\subsubsection{Error handling}

Due to the randomness and uncertainty of the responses of LLM~\cite{xiong2023can}, their behavior is significantly different from traditionally manually designed evolutionary search operators, so the LLM may generate responses that deviate from the expected format, resulting in incorrect recognition by the framework. To avoid such situations, we rigorously validate the generated text responses and initiate a new context learning process if necessary. 
To ensure that the search capability of the LLM is not influenced by historical information, we adopt a single-round polling strategy to eliminate the interference of historical information, allowing the model to always focus on the current prompt.

\section{Experiment}

\subsection{Test instances}
The experiments were conducted on widely used UF~\cite{li2008multiobjective} and ZDT~\cite{zitzler2000comparison} test instances, which possess different problem characteristics and various shapes of PF (Pareto Front) and PS (Pareto Set). Meanwhile, the UF instances collected data on multi-objective problems from various aspects of real life, reflecting the generalization performance of this framework without the need for excessive involvement of experts.

\subsection{Baseline algorithms}
We compared NSGA-II, NSGA-II-ARSBX ~\cite{Arsbx}, MOEA/D~\cite{zhang2007moea}, NSGA-III~\cite{deb2014nsgaiii}, MOEA/D-DRA~\cite{zhang2009moea}, MOEA/D-DQN~\cite{zhang2020moead_dqn} and our framework. Among them, NSGAII and NSGA-II-ARSBX use the same algorithmic framework as our framework, with the main difference lying in their reproductive strategy. All algorithm implementations are based on PlatEMO~\cite{tian2017platemo}.

\subsection{Experimental Settings}
Our algorithm settings are as follows:

\begin{itemize}

    \item Population number $N$: 100;

    \item Maximum evaluation quantity $N_{max}$: 10000;

    \item Prompts to enter the number of individuals $l$: 5;

    \item Number of output items $s$: 3;

    \item Decision threshold $\delta$: 0.1;

    \item Binary cross distribution index $\sigma$: 20;

    \item Mutation score Index $\tau$: 1;

    \item Polynomial mutation distribution index $\chi$: 20;
\end{itemize}

Unmentioned Settings on the algorithm are the same as in the original paper.

\subsection{Results}

\begin{figure*}[!t]
    \centering
    \subfloat[ZDT1]{\includegraphics[width=0.23\linewidth]{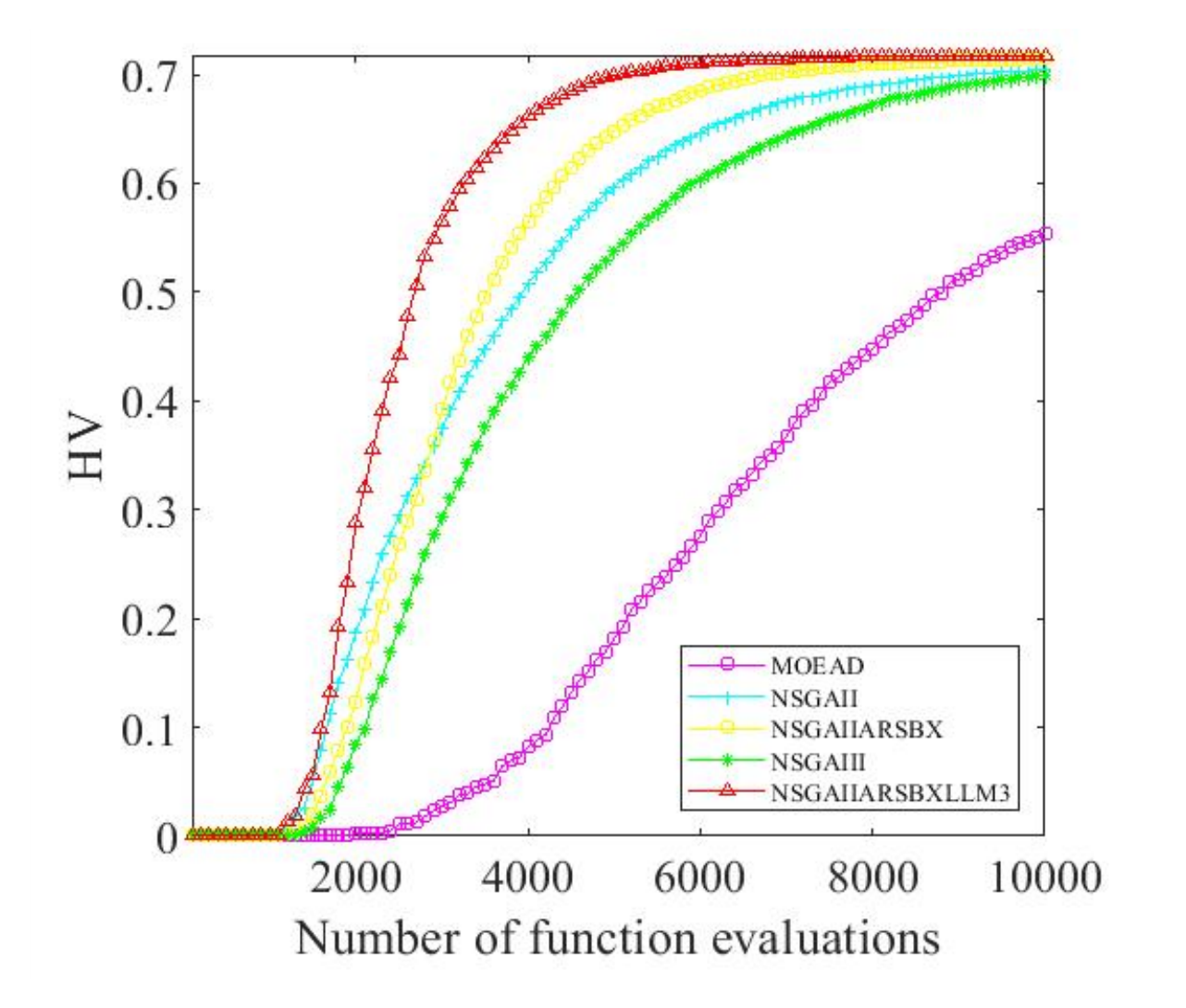}}
    \subfloat[ZDT2]{\includegraphics[width=0.23\linewidth]{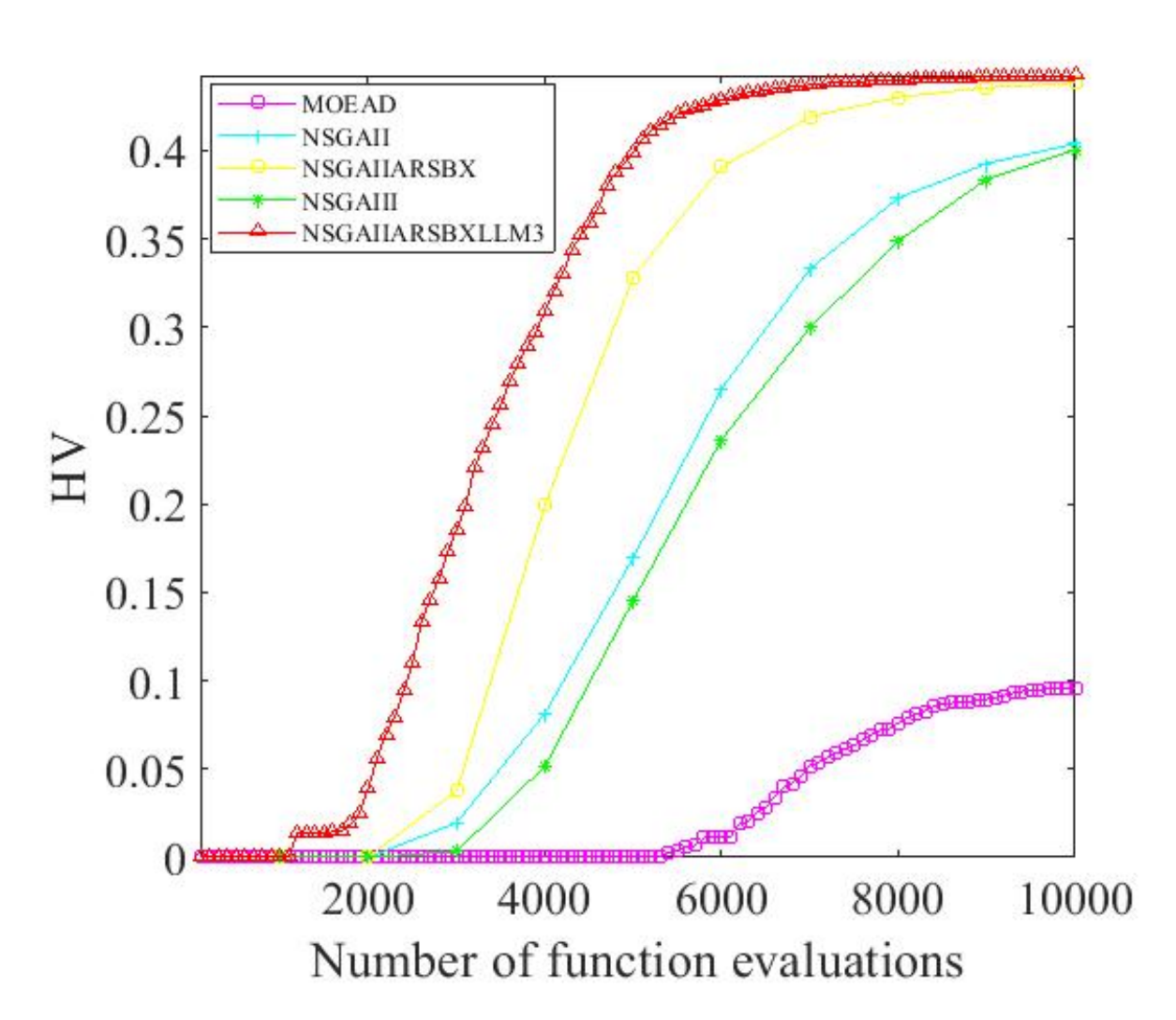}}
    \subfloat[UF1]{\includegraphics[width=0.23\linewidth]{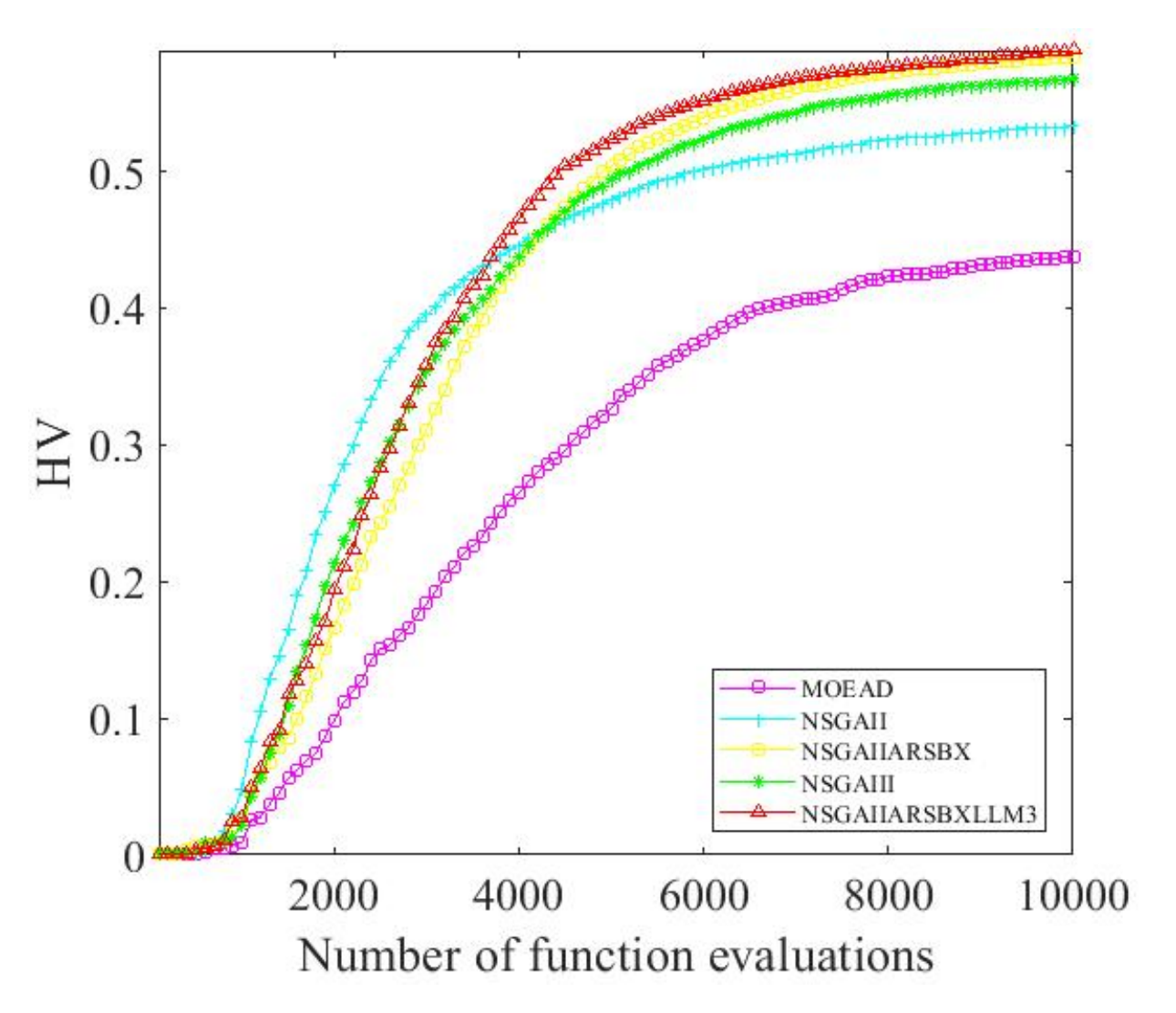}}
    \subfloat[UF2]{\includegraphics[width=0.23\linewidth]{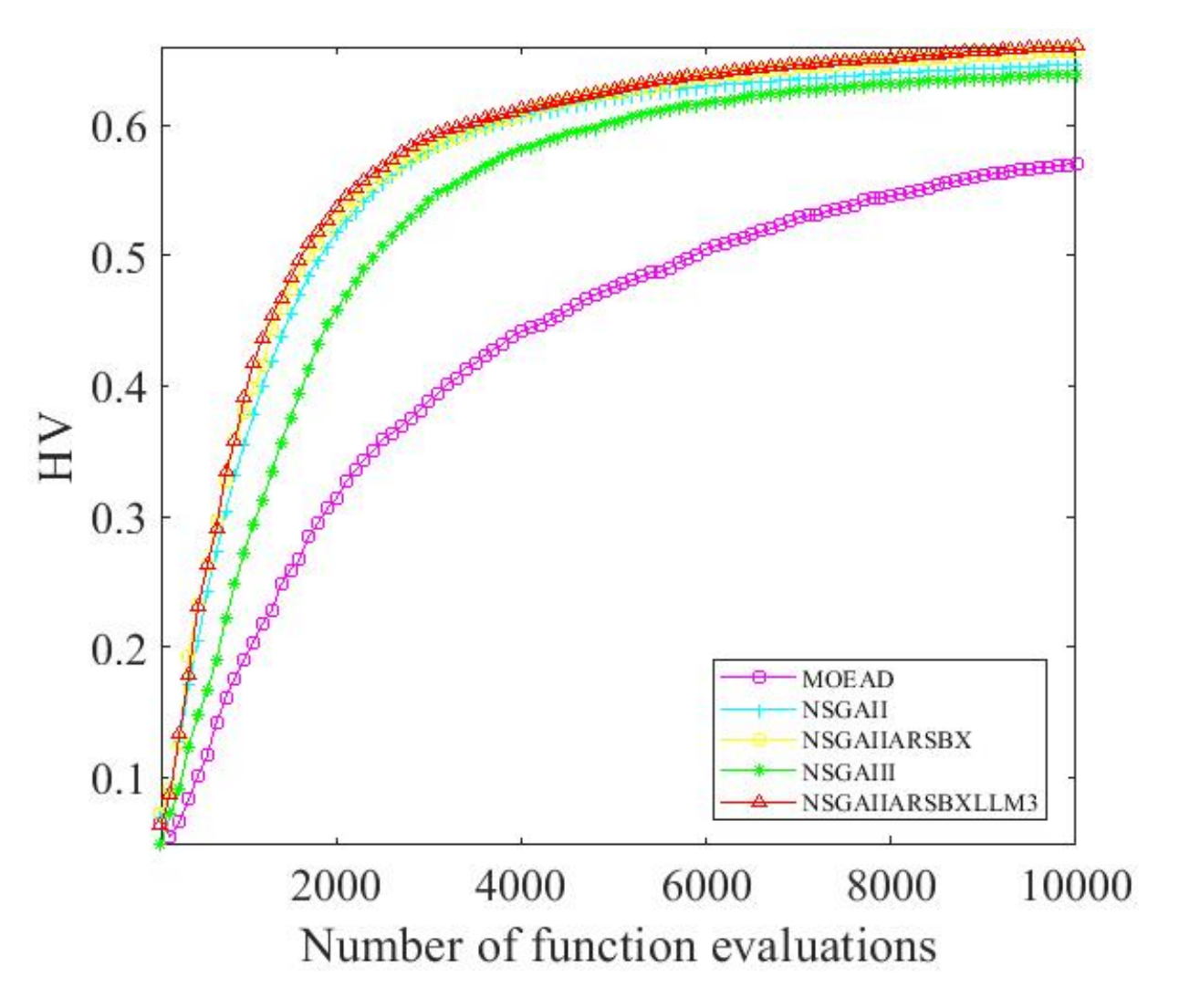}}
    \caption{HV values obtained by different multi-objective optimization algorithms.}
    \label{fig:convergence}
\end{figure*}

Figure \ref{fig:convergence} shows the convergence curves of the hypervolume (HV) values over the number of evaluations for four test instances. In the two ZDT test instances, the convergence speed of our framework is the fastest, outperforming other algorithms in terms of performance. In the UF test instances related to real-world problems, our framework also converges relatively quickly and achieves better solutions. The results indicate that our framework not only has strong competitiveness and robustness but also effectively accelerates the convergence speed and optimizes the solution set. Taking UF1 as an example, although the convergence speed of our algorithm is slightly slower than that of other algorithms in the initial stage, as the number of iterations increases, the convergence speed gradually accelerates, ultimately yielding a solution set that is superior to those obtained by other algorithms.

\begin{table*}[ht]
\centering
\caption{Compare the results of ZDT and UF examples calculated by HV.}\label{table:hv}
\resizebox{\textwidth}{!}{%
\begin{tabular}{ccccccc}
\hline
\hline
Problem    & MOEA/D          & MOEA/D-DQN          & NSGA-II          & NSGA-II-ARSBX        & NSGA-III            & NSGA-II-LLM                         \\
\hline
ZDT1     & 5.5441e-1 (6.78e-2) -                     & 6.8687e-1 (2.79e-2) -                        & 6.4219e-1 (2.03e-1) -                       & 7.1420e-1 (8.46e-4) -                       & 6.9668e-1 (4.84e-3) -                       & {\color[HTML]{FF0000} 7.1777e-1 (5.50e-4) }                       \\
ZDT2       & 1.0568e-1 (4.04e-2) -       & 4.2583e-1 (1.40e-2) -                        & 4.1981e-1 (1.15e-2) -                        & 4.3830e-1 (1.40e-3) -                        & 4.0955e-1 (1.15e-2) -                        & {\color[HTML]{FF0000} 4.4244e-1 (6.09e-4) }               \\
ZDT3      & 5.5275e-1 (5.49e-2) -       & 5.8643e-1 (1.66e-2) -                        & {\color[HTML]{FF0000} 6.0199e-1 (2.79e-2)  +}                        & 5.9654e-1 (6.57e-4) -                        & 5.8757e-1 (3.20e-3) -        & 5.9851e-1 (3.56e-4)             \\
ZDT4   & 1.6001e-1 9.45e-2) +  & 3.0175e-1 (1.56e-1) + & {\color[HTML]{FF0000} 5.2384e-1(1.56e-1) +} & 2.5449e-2(8.05e-2) = & 2.0294e-1(2.14e-1) = & 9.9391e-2(2.24e-1) \\
ZDT6   & 2.7401e-1(4.03e-2) - & {\color[HTML]{FF0000} 3.8879e-1(2.45e-4) +} & 3.2003e-1(3.95e-2) - & 3.8801e-1(1.99e-4) = & 2.2030e-1(5.05e-2) - & 3.8812e-1(1.98e-4)  \\
UF1    & 4.2971e-1(4.23e-2) - & 4.8881e-1(7.92e-2) - & 5.4658e-1(4.74e-2) - & 5.8548e-1(9.99e-3) = & 5.6095e-1(3.79e-2) = & {\color[HTML]{FF0000} 5.8760e-1(5.37e-3) } \\
UF2    & 5.5986e-1(4.67e-2) - & 6.5250e-1(1.22e-2) = & 6.4400e-1(9.42e-3) - & 6.5703e-1(5.95e-3) = & 6.4166e-1(6.40e-3) - & {\color[HTML]{FF0000} 6.5787e-1(6.13e-3) }\\
UF3    & 3.2073e-1(2.32e-2) - & 3.7056e-1(1.58e-2) - & 2.8297e-1(4.69e-2) - & 4.5375e-1(5.49e-2) = & 2.5261e-1(4.53e-2) - & {\color[HTML]{FF0000} 4.6415e-1(2.44e-2) }\\
UF4    & 2.7356e-1(8.03e-3) - & 3.4066e-1(1.23e-2) - & 3.4289e-1(6.12e-3) - & 3.4839e-1(2.86e-3) - & 3.3845e-1(3.89e-3) - & {\color[HTML]{FF0000} 3.5873e-1(3.54e-3) }\\
UF5    & 0.0000e+0(0.00e+0) - & 8.2041e-3(1.39e-2) - & 3.8367e-2(5.47e-2) - & 2.7570e-2(3.82e-2) - & 2.7416e-2(4.25e-2) - & {\color[HTML]{FF0000} 1.0782e-1(5.95e-2) } \\
UF6    & 1.2113e-1(6.42e-2) = & {\color[HTML]{FF0000} 1.4500e-1(9.95e-2) =} & 9.7397e-2(7.54e-2) = & 1.1481e-1(6.52e-2) = & 8.3580e-2(7.02e-2) = & 1.2278e-1(6.51e-2)  \\
UF7    & 2.1599e-1(6.18e-2) - & 2.9221e-1(1.41e-1) - & 3.7465e-1(9.21e-2) - & 5.0681e-1(9.75e-3) = & 4.0941e-1(9.08e-2) - & {\color[HTML]{FF0000} 5.1215e-1(3.58e-3) } \\
UF8    & 1.5169e-1(5.62e-2) - & 2.1856e-1(5.82e-2) = & 2.8090e-1(2.88e-2) + & 1.8354e-1(3.85e-2) = & {\color[HTML]{FF0000} 2.8998e-1(4.64e-2) +} & 2.1990e-1(4.00e-2)   \\
UF9    & 2.5698e-1(4.05e-2) = & 2.7526e-1(5.76e-2) = & {\color[HTML]{FF0000} 3.8011e-1(5.04e-2) +} & 2.5736e-1(3.93e-2) = & 3.1900e-1(5.31e-2) + & 2.5167e-1(5.67e-2)  \\
UF10   & {\color[HTML]{FF0000} 5.1911e-2(2.67e-2) +} & 6.1359e-3(1.29e-2) = & 1.7998e-3(5.69e-3) = & 0.0000e+0(0.00e+0) = &3.5134e-3(6.68e-3) = & 2.1857e-2(6.55e-2)  \\
\hline
\multicolumn{1}{c}{+/-/=} & 2/11/2     & 2/8/5      & 4/9/2    & 0/5/10 & 2/9/4 &    \\
\hline
\hline
\end{tabular}%
}
\end{table*}

\begin{table*}[ht]
\centering
\caption{The results of ZDT and UF examples are compared by IGD.}\label{table:igd}
\resizebox{\textwidth}{!}{%
\begin{tabular}{ccccccc}
\hline
\hline
Problem    & MOEA/D          & MOEA/D-DQN          & NSGA-II          & NSGA-II-ARSBX        & NSGA-III            & NSGA-II-LLM           \\
\hline
ZDT1       & 1.7796e-1(8.73e-2) - & 2.7122e-2(2.10e-2) - & 7.6042e-2(2.01e-1) - & 7.0553e-3(5.43e-4) - & 1.9184e-2(3.50e-3) - & {\color[HTML]{FF0000} 5.0831e-3(2.64e-4) }  \\
ZDT2       & 5.3206e-1(1.07e-10 - & 1.4418e-2(9.17e-3) - & 1.9256e-2(7.51e-3) - & 7.1560e-3(6.33e-4) - & 2.7104e-2(8.00e-3) - & {\color[HTML]{FF0000} 5.1734e-3(3.04e-4) }  \\
ZDT3       & 1.5641e-1(6.51e-2) - & 2.5088e-2(2.66e-2) - & 1.2816e-2(1.01e-2) - & 7.1529e-3(7.03e-4) - & 1.6483e-2(3.79e-3) - & {\color[HTML]{FF0000} 5.5914e-3(2.95e-4) }\\
ZDT4       & 5.2553e-1(1.47e-1) + & 3.8006e-1(2.01e-1) + & {\color[HTML]{FF0000} 2.0151e-1(1.73e-1) +} & 1.5667e+0(5.51e-1) - & 5.8821e-1(3.07e-1) = & 1.0624e+0(7.07e-1) \\
ZDT6       & 1.0572e-1(7.45e-2) - & {\color[HTML]{FF0000} 3.1436e-3(3.69e-5) +} & 5.4529e-2(3.23e-2) - & 3.8839e-3(1.87e-4) = & 1.4733e-1(5.59e-2) - & 3.7911e-3(2.05e-4) \\
UF1        & 2.7868e-1(8.76e-2) - & 1.9899e-1(9.86e-2) - & 1.3670e-1(4.19e-2) - & 1.0229e-1(8.22e-3) = & 1.1551e-1(3.27e-2) = & {\color[HTML]{FF0000} 9.7783e-2(5.45e-3) }\\
UF2        & 2.2149e-1(1.02e-1) - & 6.1093e-2(1.79e-2) = & 6.5876e-2(1.01e-2) - & 5.4424e-2(5.60e-3) = & 6.9106e-2(1.20e-2) - & {\color[HTML]{FF0000} 5.2895e-2(5.81e-3) }\\
UF3        & 3.3172e-1(1.42e-2) - & 2.6718e-1(2.85e-2) - & 3.4155e-1(5.11e-2) - & 2.1573e-1(5.22e-2) = & 3.8719e-1(4.60e-2) - & {\color[HTML]{FF0000} 2.0717e-1(1.85e-2) }  \\
UF4        & 1.2276e-1(7.70e-3) - & 7.8090e-2(7.94e-3) - & 7.4799e-2(4.62e-3) - & 7.1421e-2(2.29e-3) - & 7.7367e-2(1.93e-3) - & {\color[HTML]{FF0000} 6.7237e-2(4.21e-3) } \\
UF5        & 1.3110e+0(1.93e-1) - & 1.0280e+0(3.73e-1) - & 6.7430e-1(1.66e-1) - & 5.9636e-1(2.55e-1) - & 7.4466e-1(2.53e-1) - & {\color[HTML]{FF0000} 3.7397e-1(1.28e-1) 
 } \\
UF6        & 4.5383e-1(4.03e-2) = & 4.7645e-1(2.56e-1) = & 4.0890e-1(8.86e-2) = & 4.0815e-1(8.01e-2) = & 4.2625e-1(1.00e-1) = & {\color[HTML]{FF0000} 3.8078e-1(1.21e-1)   }    \\
UF7        & 4.8297e-1(1.12e-1) - & 3.5834e-1(2.33e-1) - & 2.0882e-1(1.31e-1) - & 5.7723e-2(7.20e-3) - & 1.6737e-1(1.39e-1) - & {\color[HTML]{FF0000} 5.1320e-2(3.27e-3)    }   \\
UF8        & 4.8364e-1(1.84e-1) - & 3.3588e-1(4.66e-2) = & {\color[HTML]{FF0000} 2.8168e-1(1.53e-2) +} & 3.4108e-1(2.29e-2) - & 4.4972e-1(7.48e-2) - & 3.1639e-1(2.27e-2)    \\
UF9        & 5.8640e-1(6.84e-2) - & 4.8219e-1(5.42e-2) = & {\color[HTML]{FF0000} 3.5232e-1(6.21e-2) +} & 4.8848e-1(4.94e-2) = & 4.8465e-1(8.90e-2) = & 5.0417e-1(5.62e-2) \\
UF10       & {\color[HTML]{FF0000} 7.1567e-1(4.60e-2) =} & 1.8184e+0(8.24e-1) = & 1.1979e+0(5.21e-1) = & 1.5879e+0(4.25e-1) - & 7.3506e-1(1.51e-1) = & 1.2037e+0(6.16e-1) \\
\hline
\multicolumn{1}{c}{+/-/=} & 1/12/2 & 2/8/5 & 3/10/2 & 0/9/6 & 0/10/5 &      \\
\hline
\hline                                     
\end{tabular}%
}
\end{table*}

The table~\ref{table:hv} and the table~\ref{table:igd} show the HV and IGD values of each algorithm (MOEA/D, MOEA/D-DQN, NSGA-II, NSGA-II-ARSBX, NSGA-III and our framework) on the ZDT and UF test instances. For each test instance, we ran it 10 times independently. All data are average values, standard deviations in parentheses, best values in red. The tables can significantly observe the auxiliary role of the LLM, and the LLM only needs a few examples to derive excellent solutions, without any artificial design and domain knowledge. Moreover, due to the rich prior knowledge of LLM, our framework shows strong performance on UF examples of real problems, which proves that the application of LLM and MOEA can assist in solving complex optimization problems in real life.

\begin{figure*}[htbp]
\centering
    \subfloat[]{\includegraphics[width=0.23\linewidth]{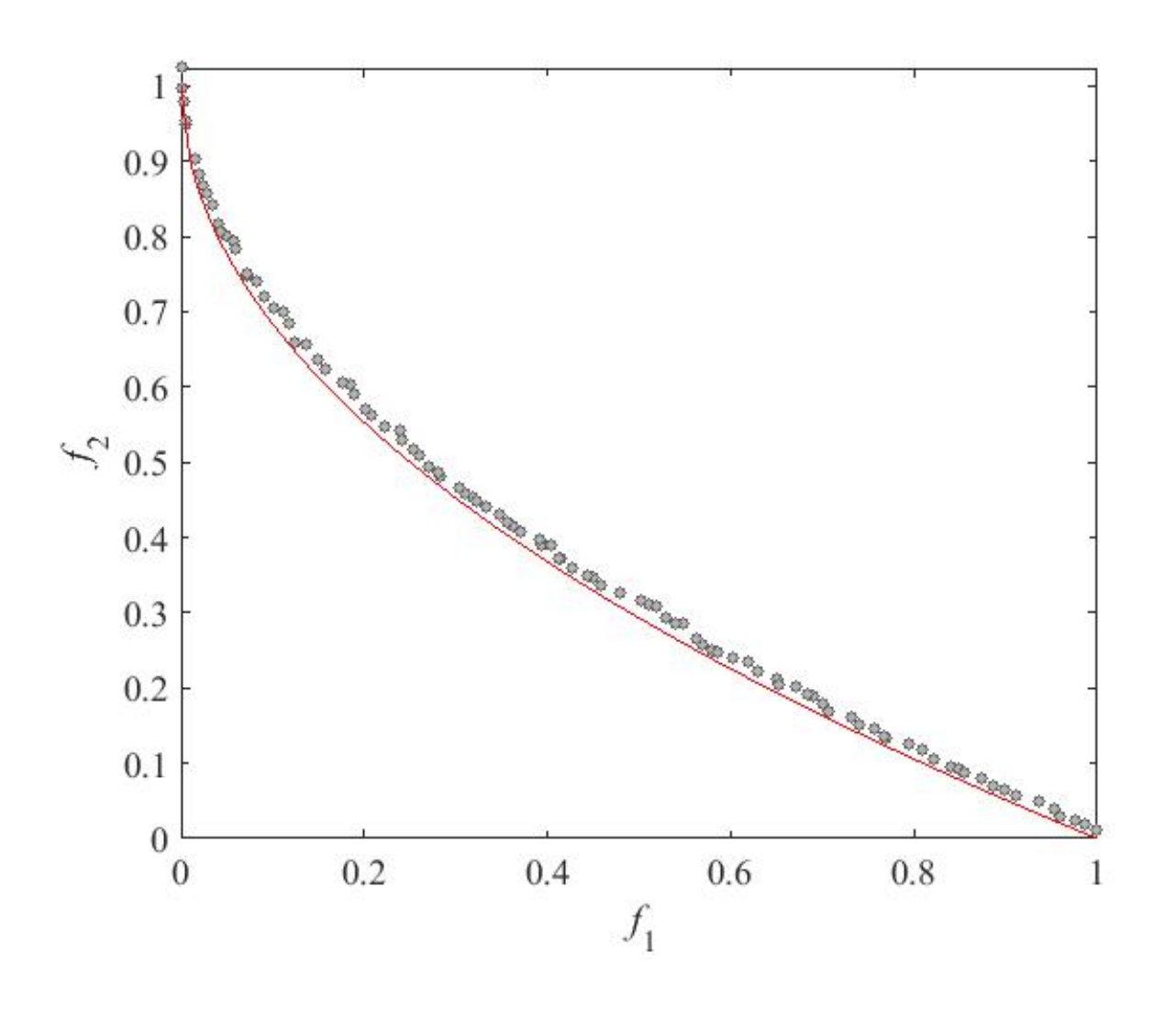}}
    \hfill
    \subfloat[]{\includegraphics[width=0.23\linewidth]{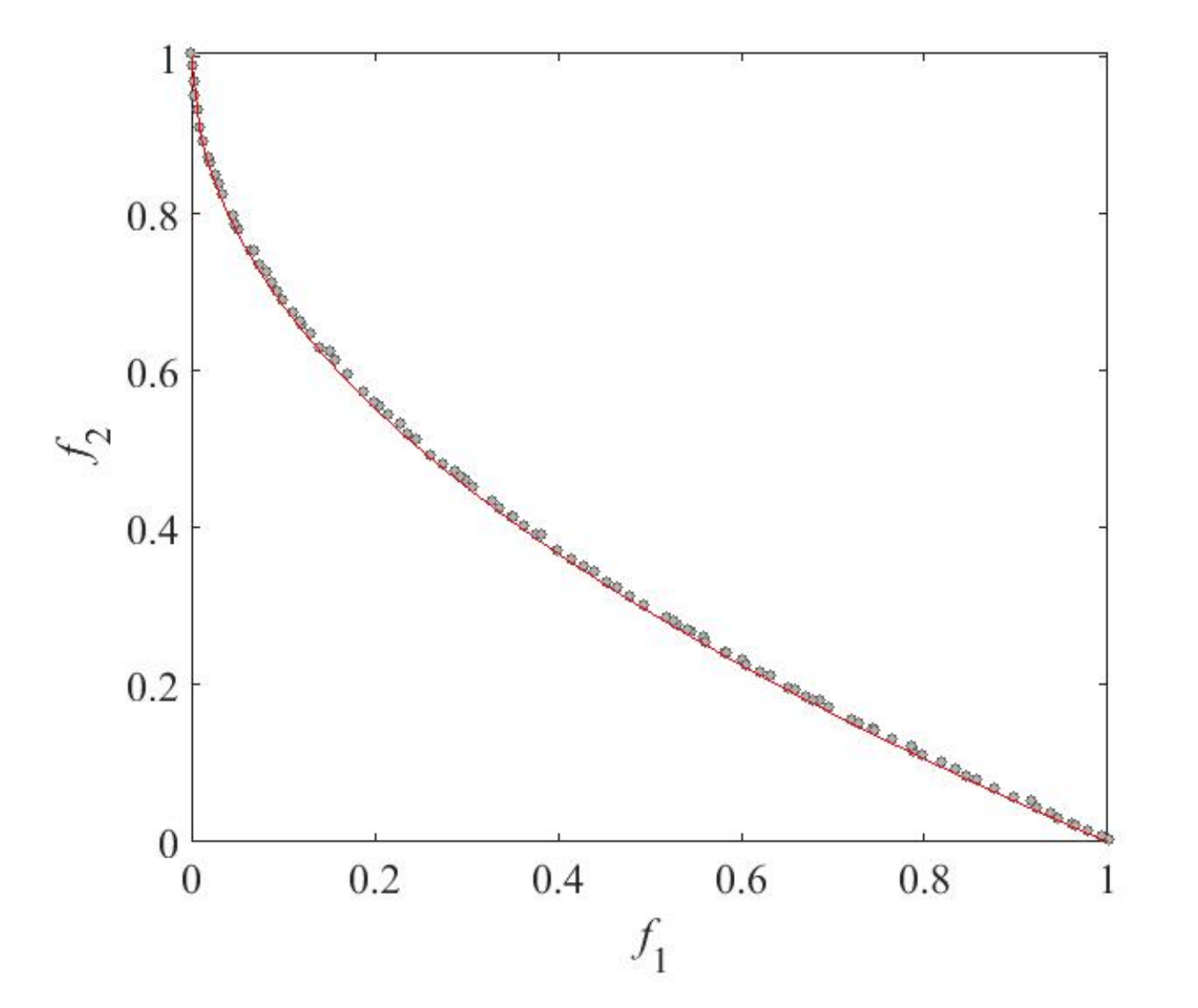}}
    \hfill
    \subfloat[]{\includegraphics[width=0.23\linewidth]{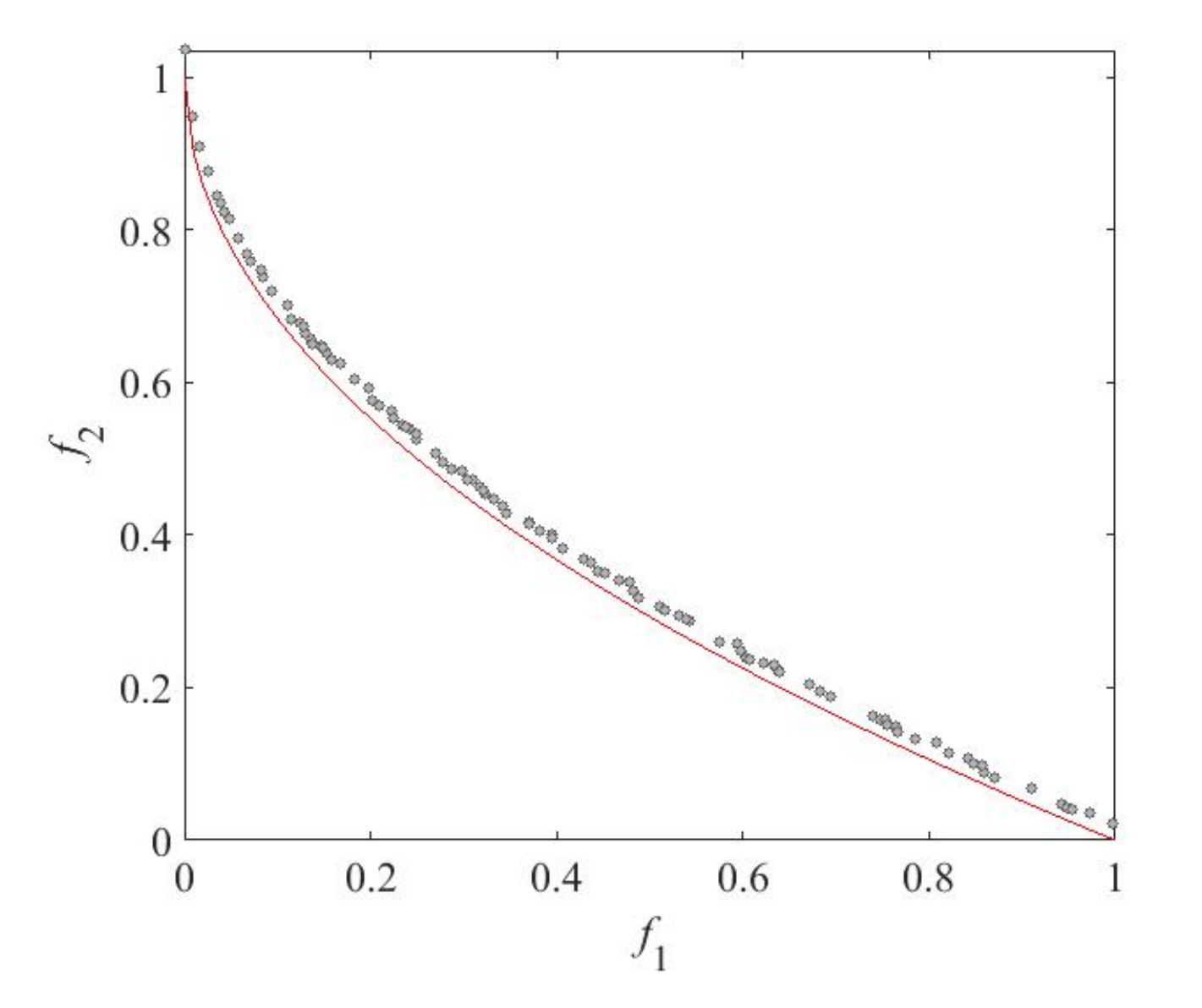}}
    \hfill
    \subfloat[]{\includegraphics[width=0.23\linewidth]{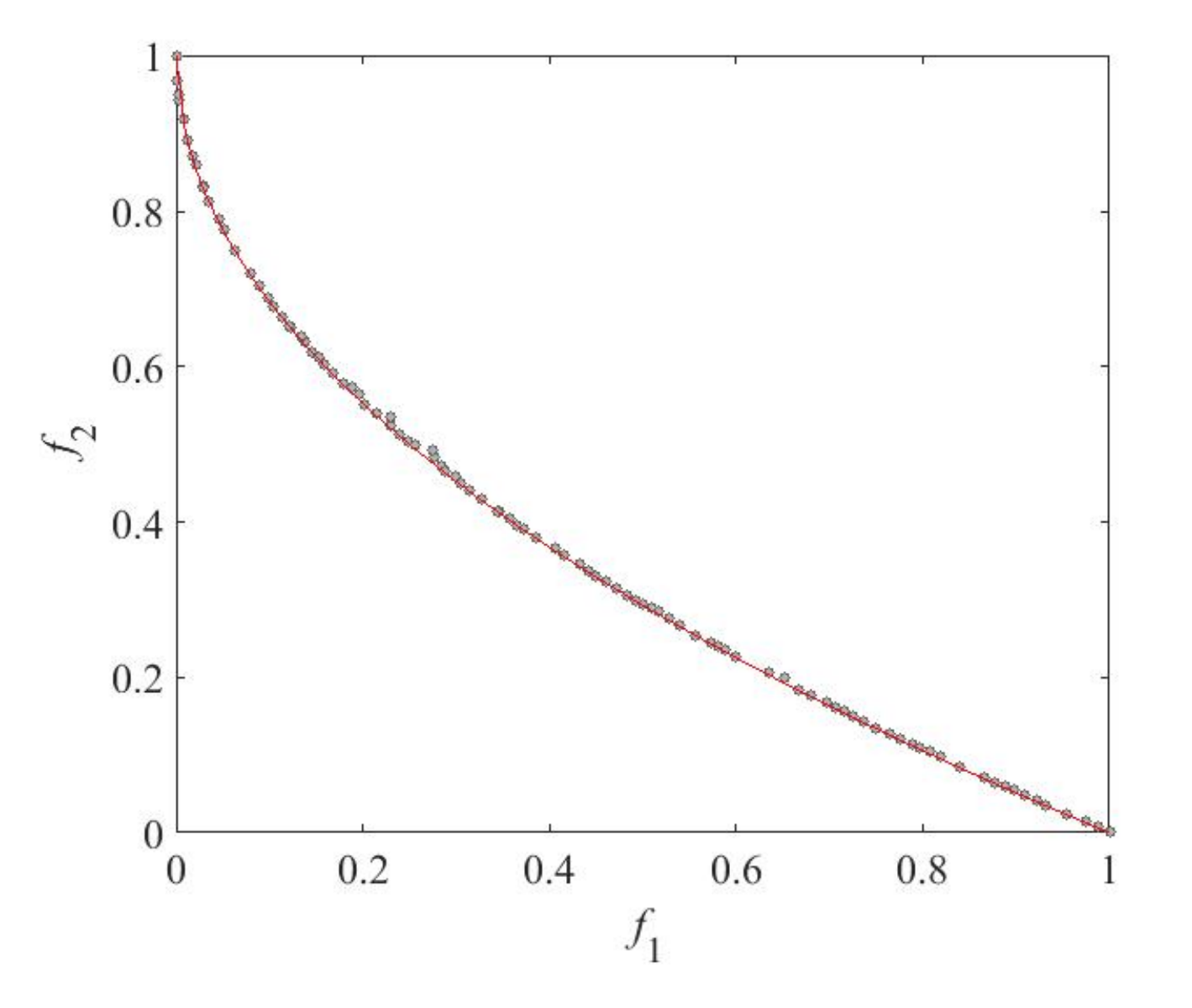}}
    \caption{Each algorithm is based on the PF surface of the ZDT1 instance: (a) NSGA-II, (b) NSGA-II-ARSBX, (c) NSGA-III, and (d) NSGA-II-LLM.}~\label{fig:pf_zdt1}
    
    \subfloat[]{\includegraphics[width=0.23\linewidth]{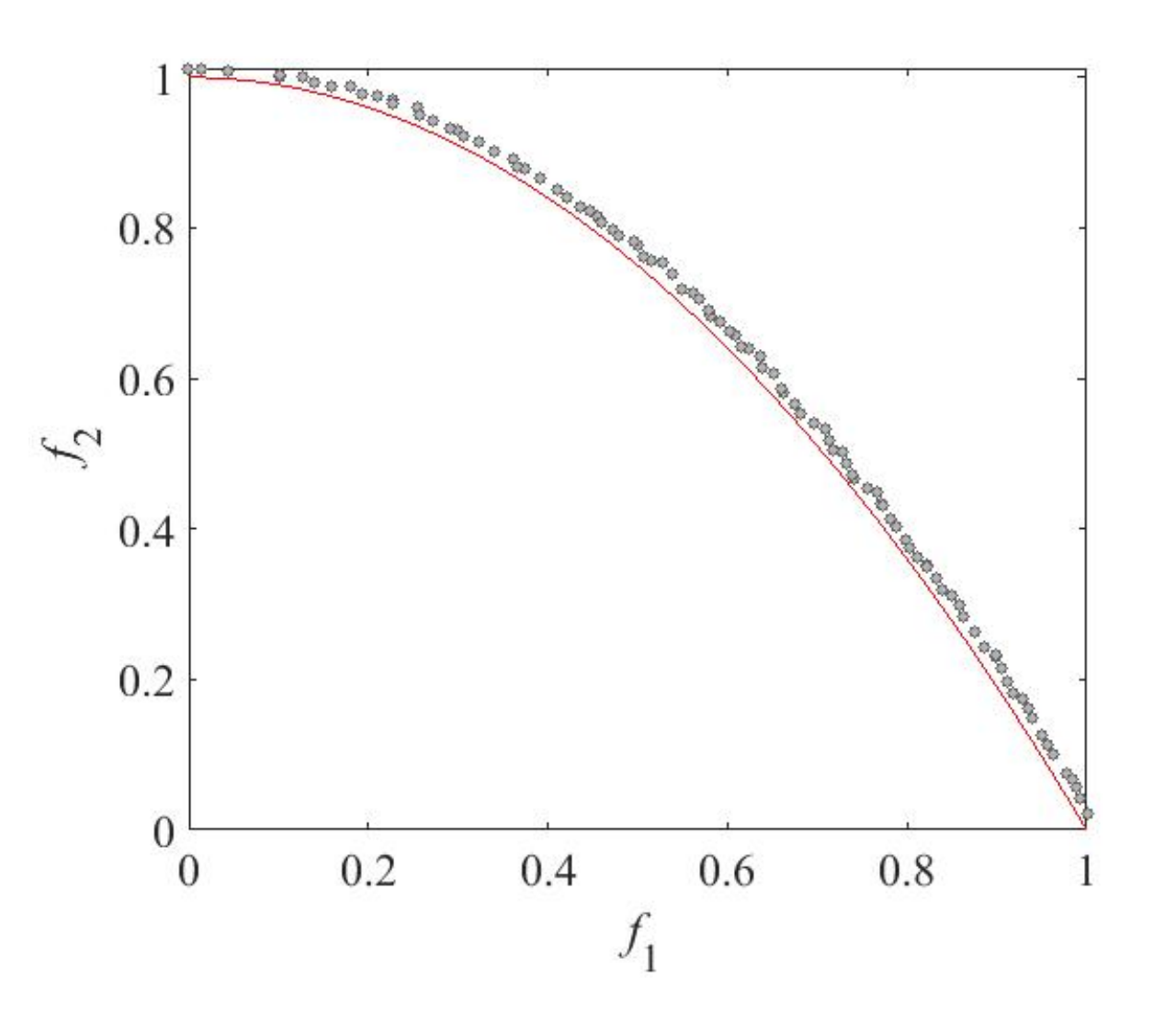}}
    \hfill
    \subfloat[]{\includegraphics[width=0.23\linewidth]{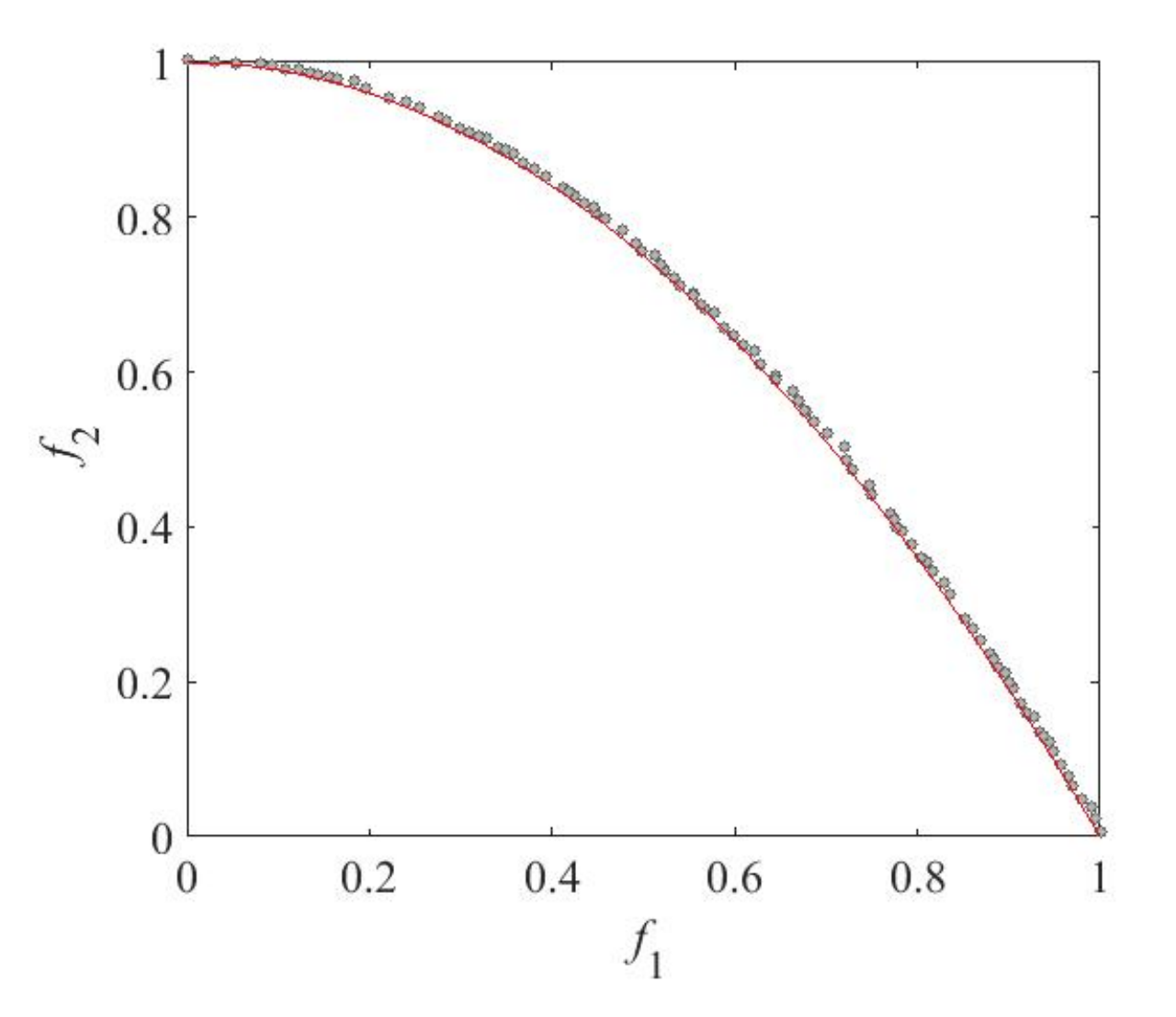}}
    \hfill
    \subfloat[]{\includegraphics[width=0.23\linewidth]{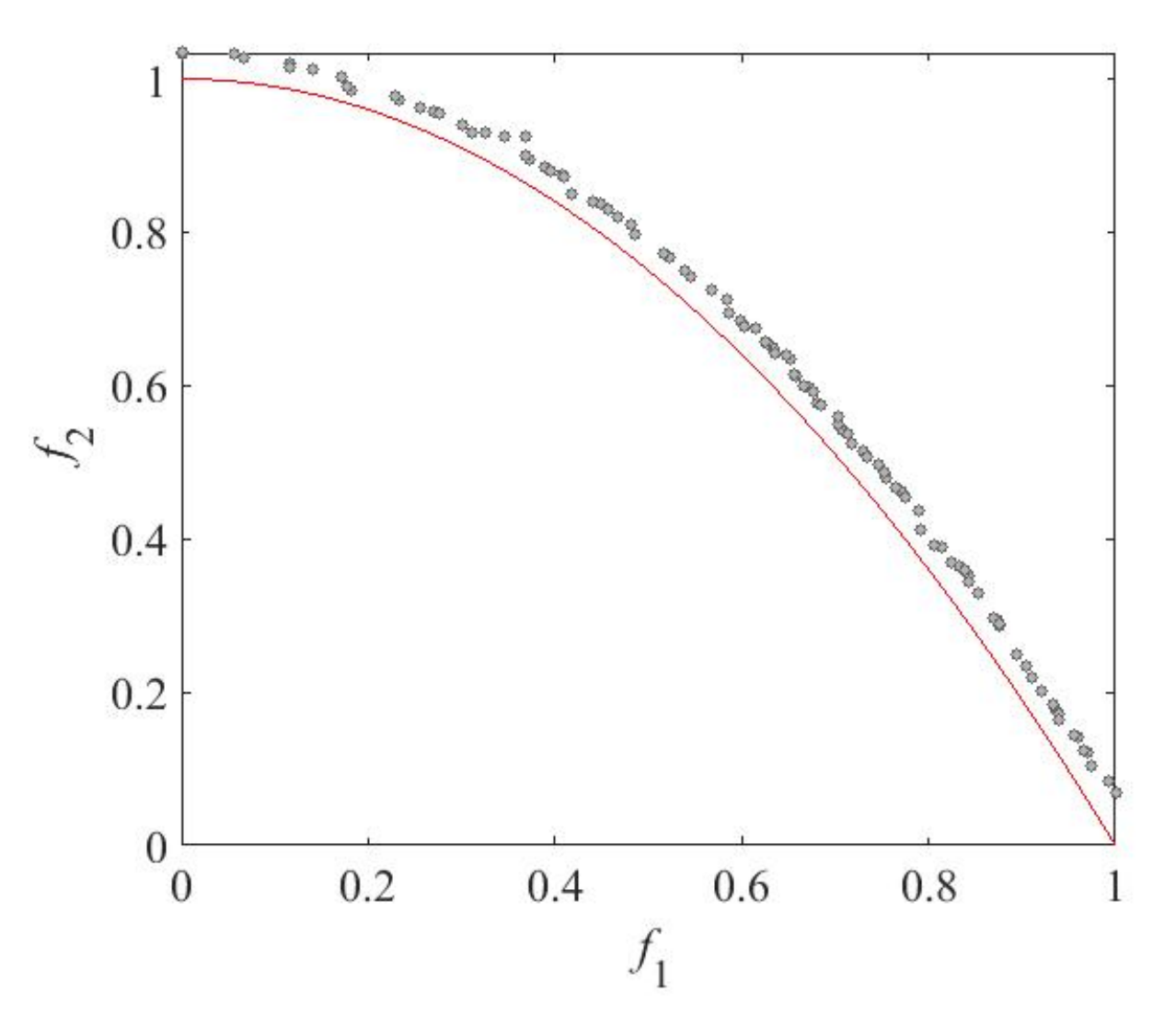}}
    \hfill
    \subfloat[]{\includegraphics[width=0.23\linewidth]{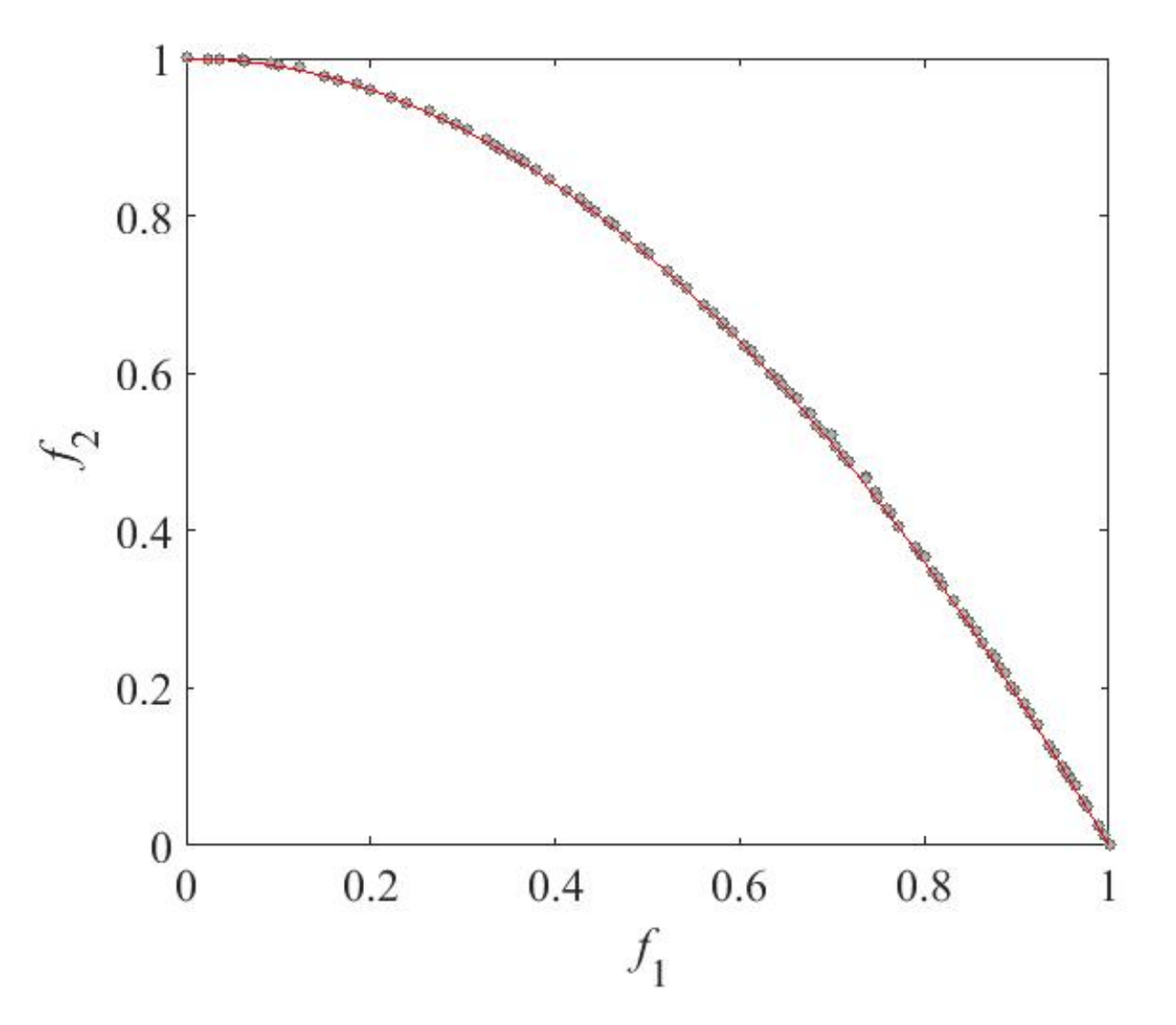}}
    \caption{Each algorithm is based on the PF surface of the ZDT2 instance: (a) NSGA-II, (b) NSGA-II-ARSBX, (c) NSGA-III, and (d) NSGA-II-LLM.}~\label{fig:pf_zdt2}
    
    \subfloat[]{\includegraphics[width=0.23\linewidth]{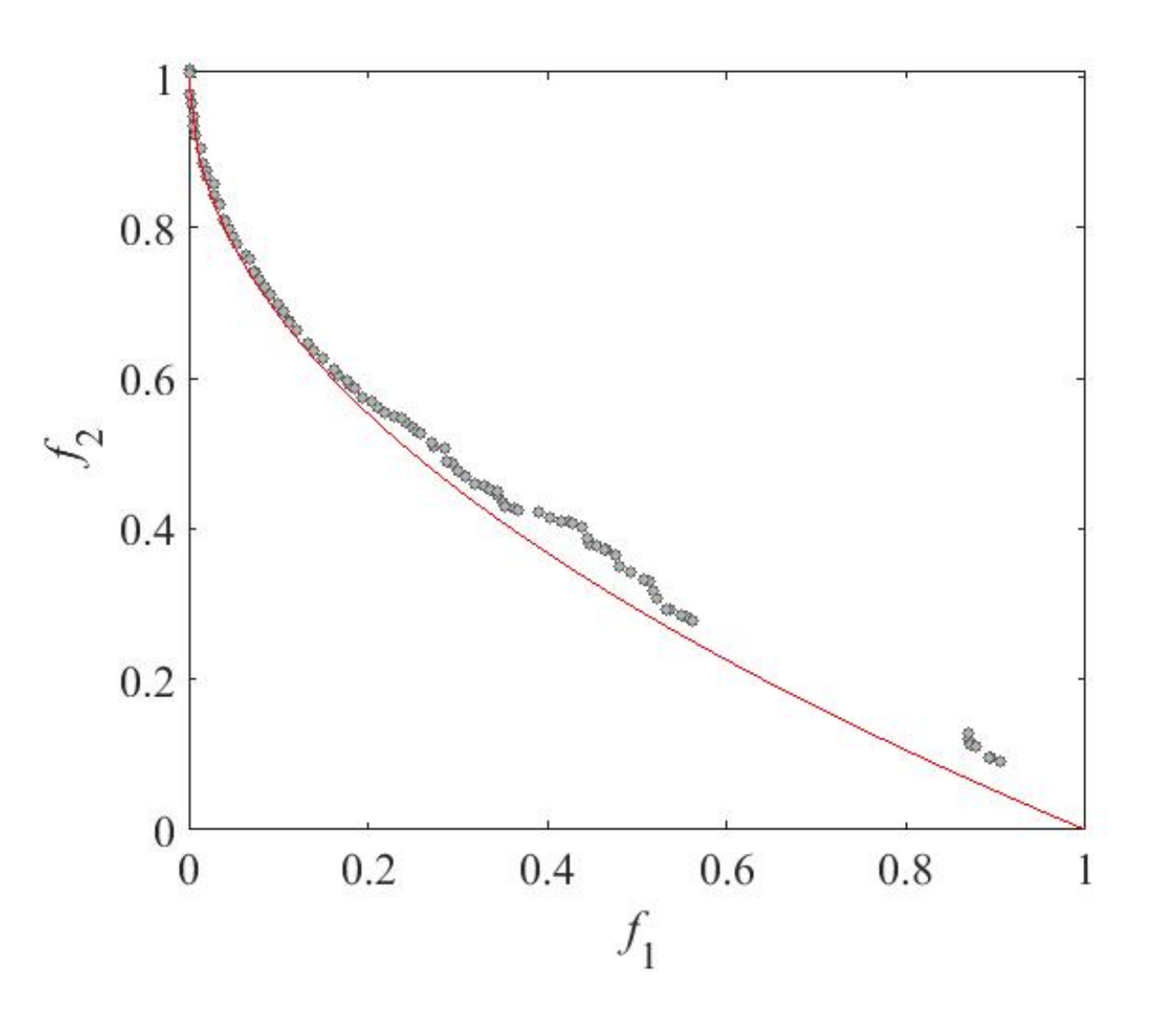}}
    \hfill
    \subfloat[]{\includegraphics[width=0.23\linewidth]{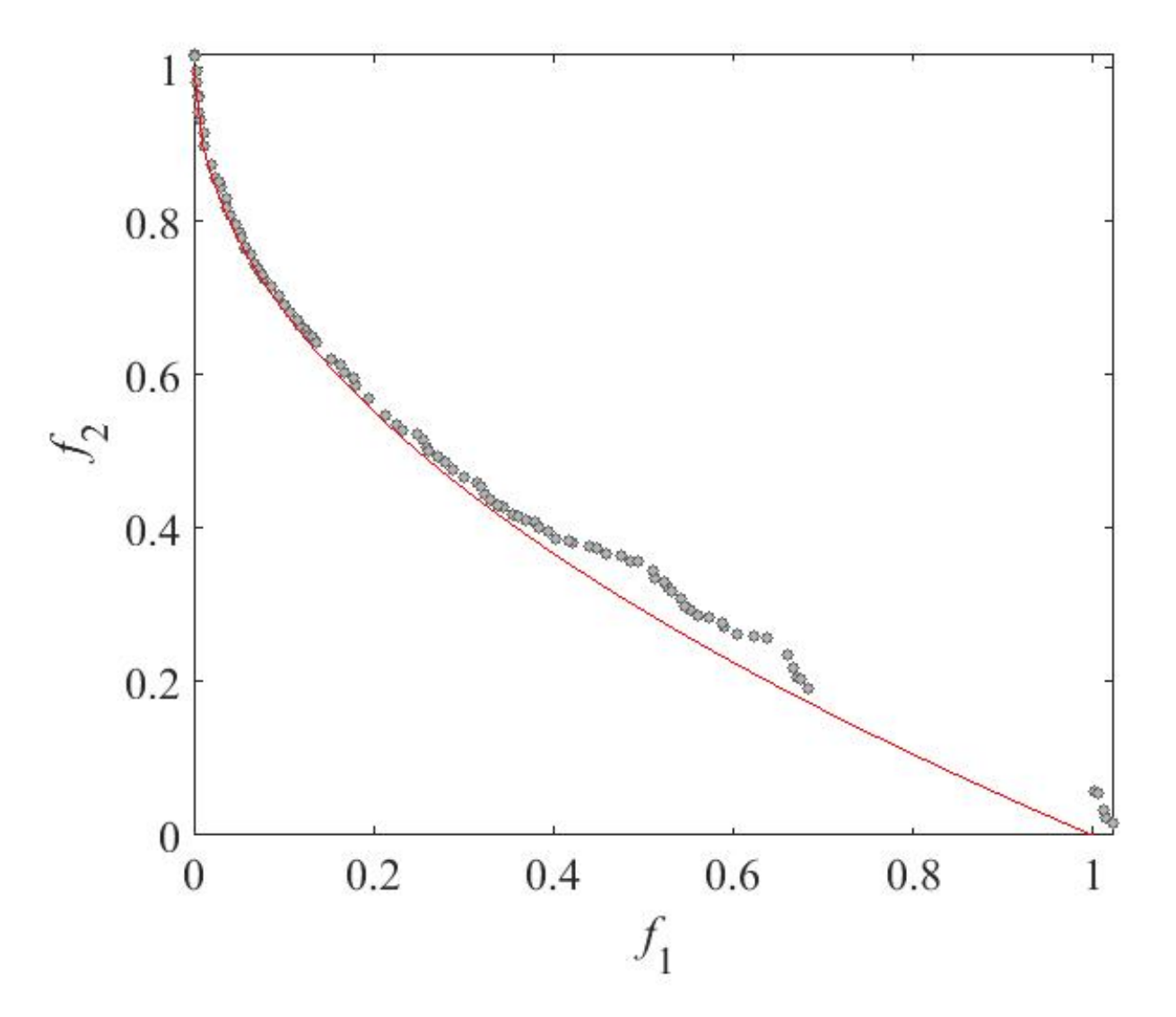}}
    \hfill
    \subfloat[]{\includegraphics[width=0.23\linewidth]{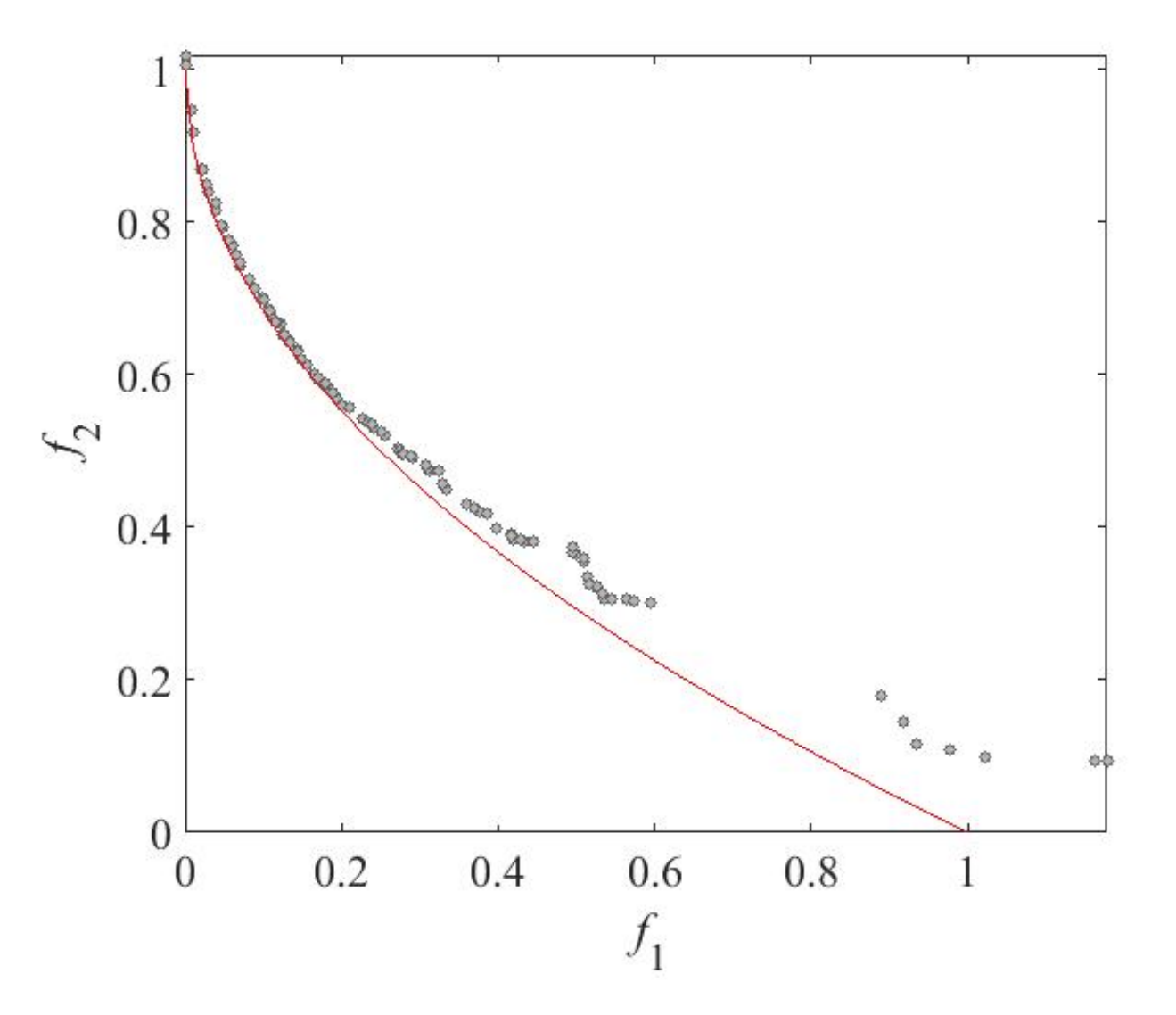}}
    \hfill
    \subfloat[]{\includegraphics[width=0.23\linewidth]{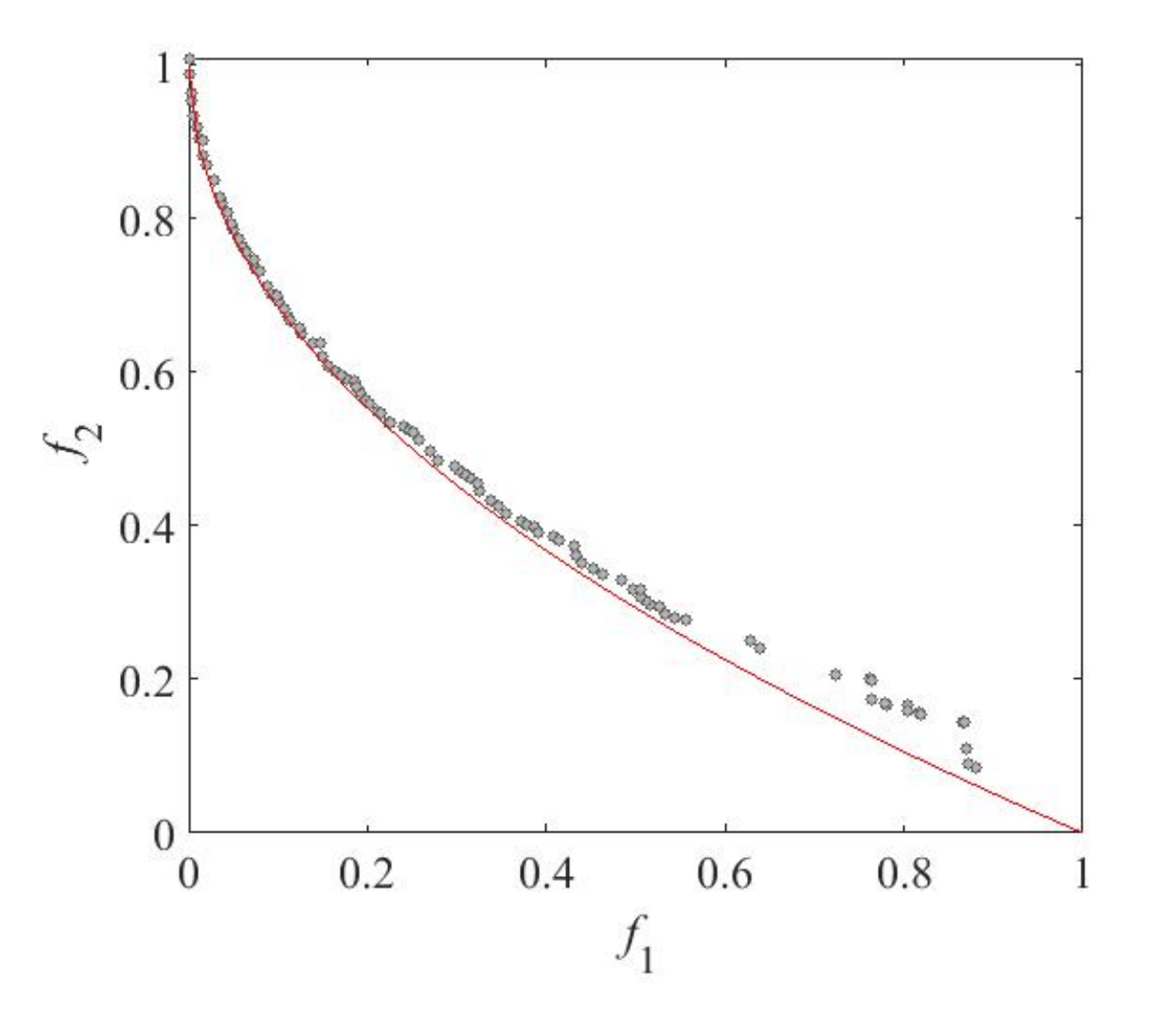}}
    \caption{Each algorithm is based on the PF surface of the UF2 instance: (a) NSGA-II, (b) NSGA-II-ARSBX, (c) NSGA-III, and (d) NSGA-II-LLM.}~\label{fig:pf_uf2}
    
    \subfloat[]{\includegraphics[width=0.23\linewidth]{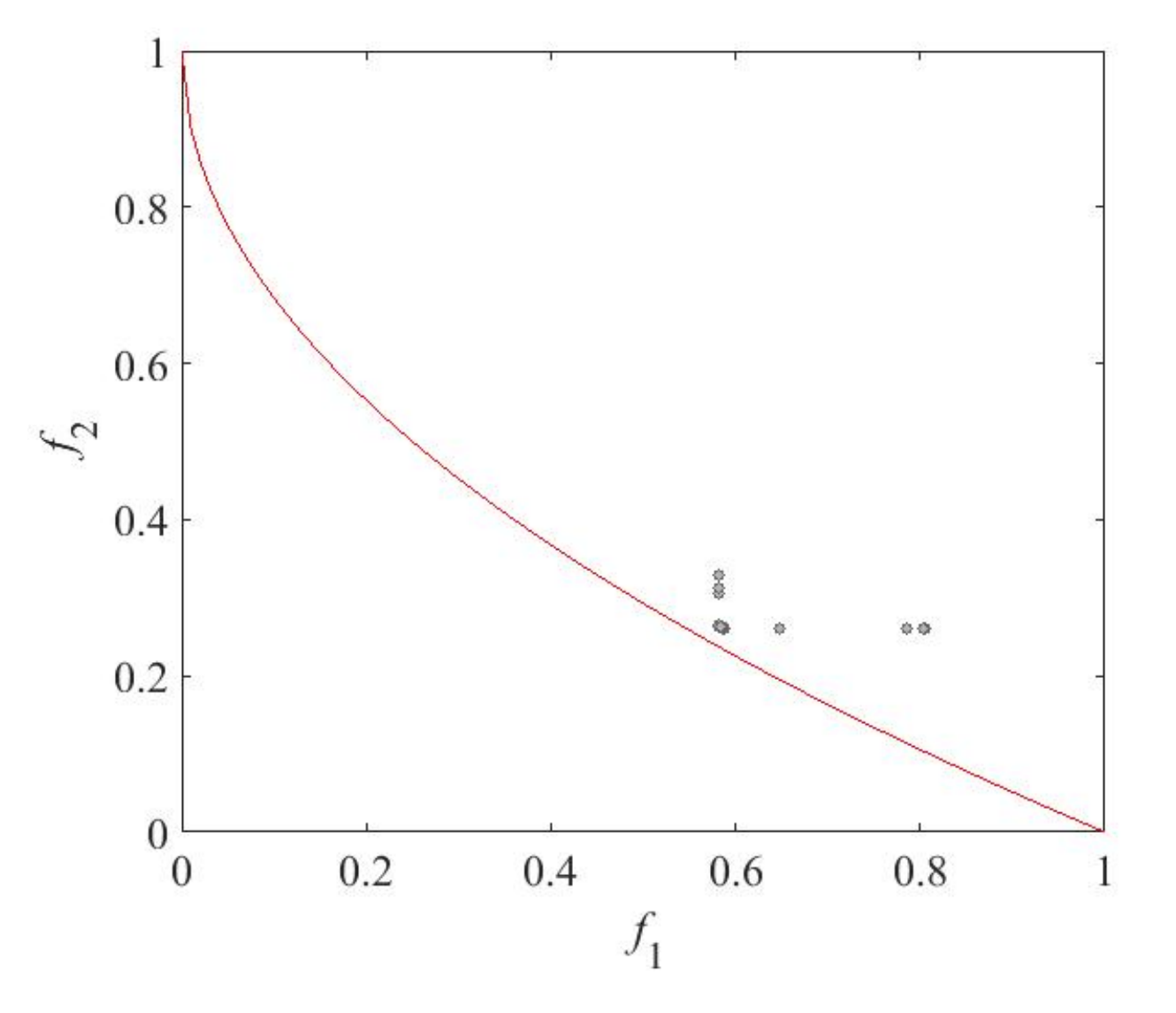}}
    \hfill
    \subfloat[]{\includegraphics[width=0.23\linewidth]{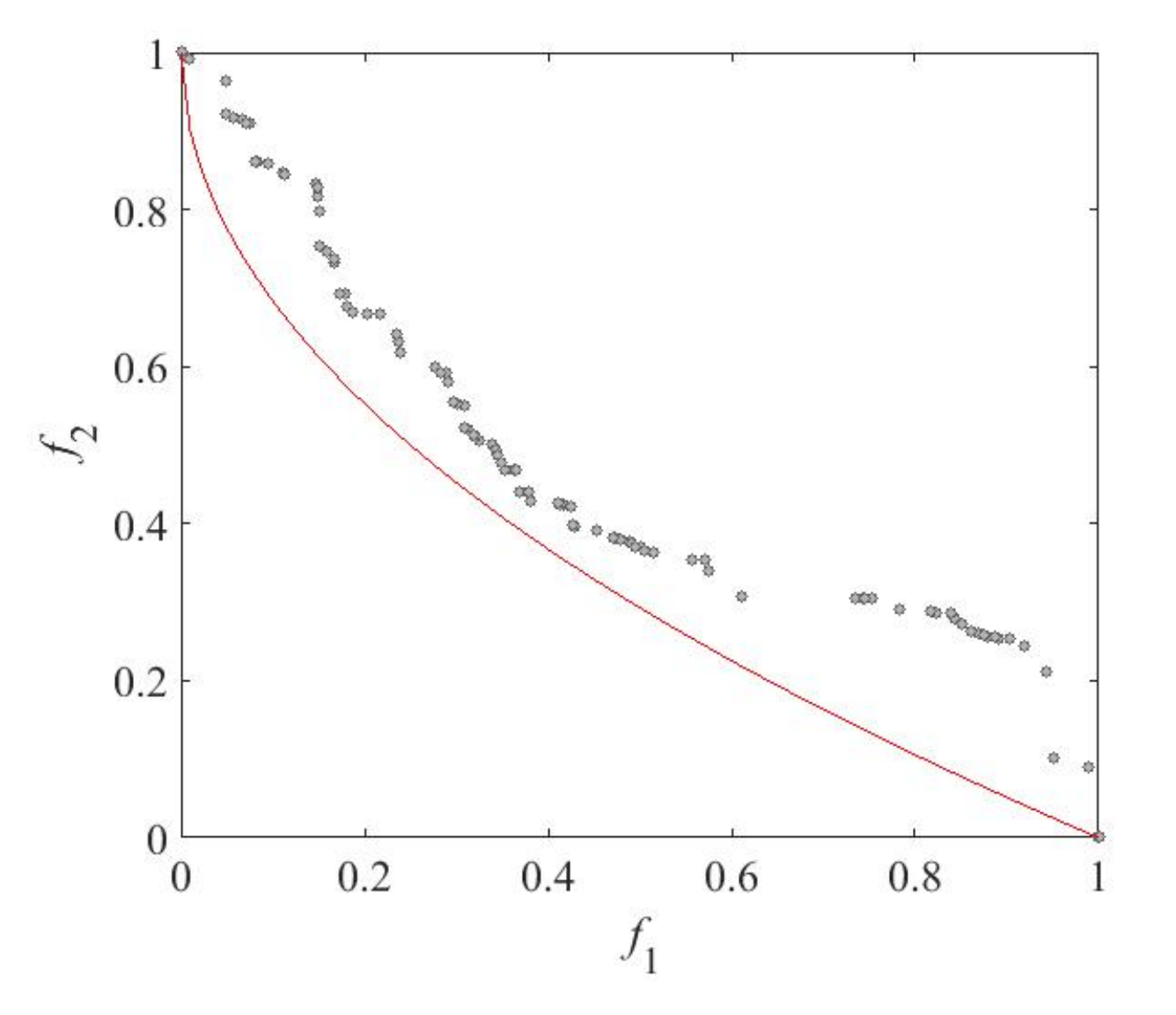}}
    \hfill
    \subfloat[]{\includegraphics[width=0.23\linewidth]{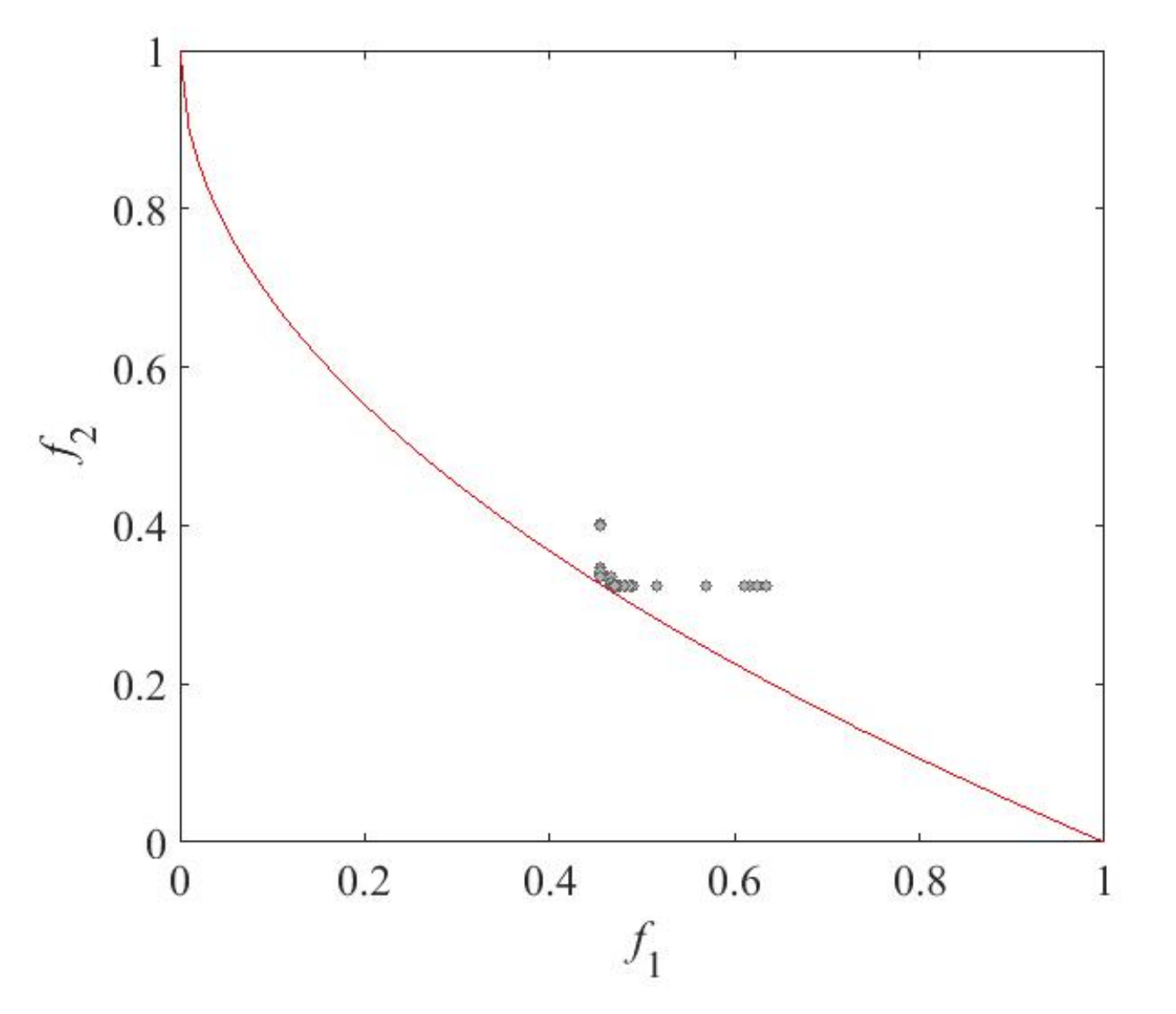}}
    \hfill
    \subfloat[]{\includegraphics[width=0.23\linewidth]{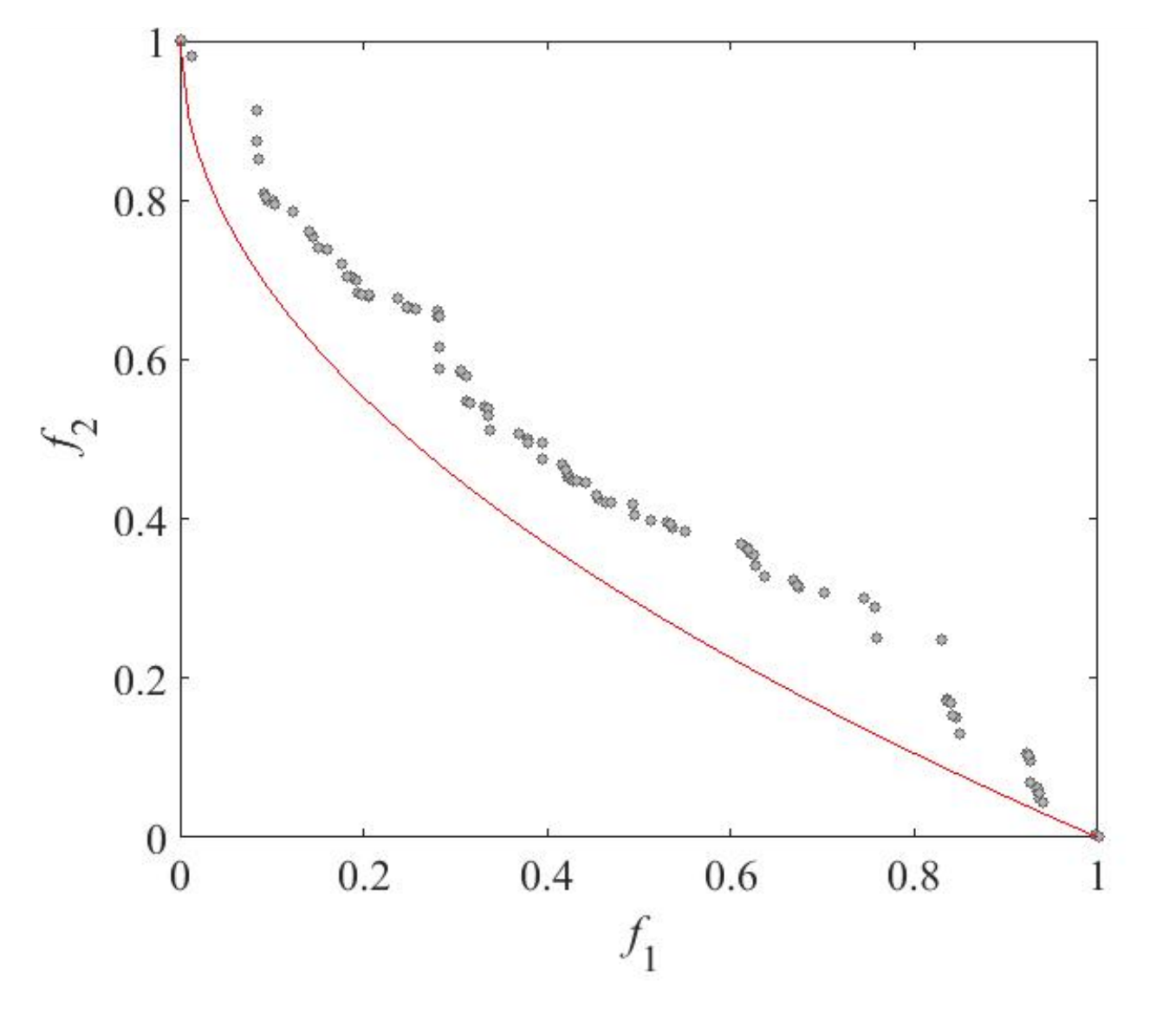}}
    \caption{Each algorithm is based on the PF surface of the UF3 instance: (a) NSGA-II, (b) NSGA-II-ARSBX, (c) NSGA-III, and (d) NSGA-II-LLM.}~\label{fig:pf_uf3}

\end{figure*}

In addition, in order to verify that the auxiliary role of the LLM can indeed help the MOEA accelerate the convergence rate of the population and enrich the diversity of the population, we present the Pareto Front (PF) surfaces after running the ZDT and UF instances.The results in Figures \ref{fig:pf_zdt1} and \ref{fig:pf_zdt2} indicate that our framework has a higher convergence rate and a better fitting effect on the ZDT instances.
For UF2 and UF3 instances in Figures\ref{fig:pf_uf2} and \ref{fig:pf_uf3}, our framework achieves a superior distribution of diversity, with solutions present in all directions, and it more accurately fits the actual PF surface. This once again strongly suggests that the introduction of LLM not only helps to accelerate convergence but also contributes to maintaining the diversity of the solutions.

\subsection{Ablation Study}
The proposed framework has produced satisfactory results in the experiment. However, the decision threshold $\delta$ requires ablation studies to verify that it can effectively help the framework reduce the number of LLM interactions and reduce costs.

\begin{figure}[hbtp]
    \centering
    \subfloat[]{\includegraphics[width=0.45\textwidth]{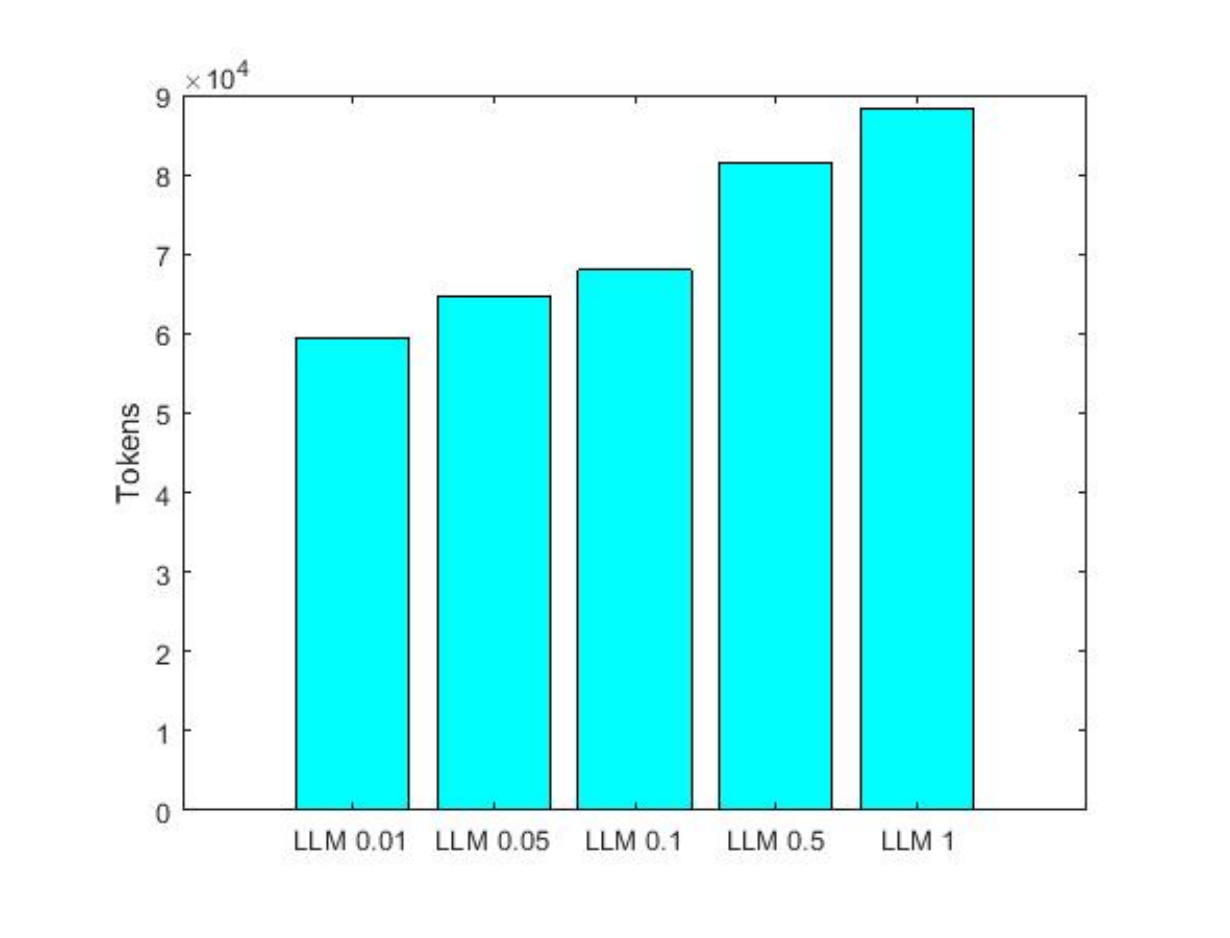}}
    \hfill
    
    \subfloat[]{\includegraphics[width=0.45\textwidth]{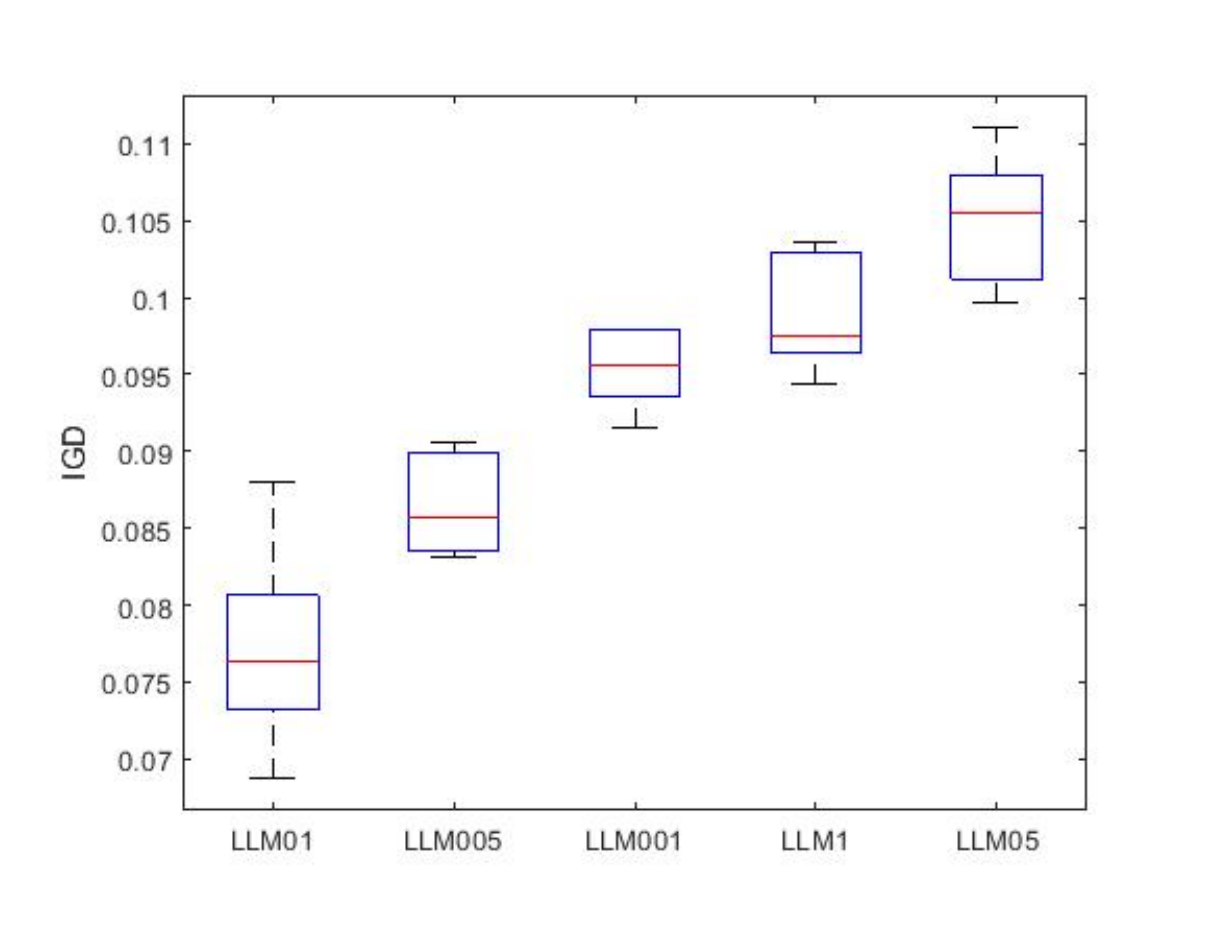}}
    \caption{(a) The number of tokens required by each decision threshold for a single run, and (b) the average IGD comparison of UF instances..}
    \label{fig:ablation}
\end{figure}

To conduct the ablation study, we tested the score range $(0,1]$ of the auxiliary evaluation function. Depending on its range, we test with input decision threshold sizes of $\{0.01,0.05,0.1, 0.5,1\}$ while keeping all other experimental settings unchanged.As shown in Figure\ref{fig:ablation}, We tested the number of tokens used in a single run of these inputs and the IGD vaules on the UF instance. 
The experimental results confirm that in the case of no significant difference, igd performs best and the token cost is small when the input is $0.1$.

\section{Future Works}

In future work, we will focus on applying large language models to multi-objective evolutionary algorithms in the following directions:
\begin{itemize}
    \item \textbf{Automatic feature extraction and problem Modeling:} Using LLM to understand real-world problems described in natural language, and automatically extract key features and constraints. Based on the problem, generate specific solutions and construct fitness functions.

    \item \textbf{Algorithm parameter optimization:} Apply LLM to predict and optimize MOEA parameters, such as population size, crossover and mutation probability. An adaptive parameter adjustment mechanism based on LLM is developed to deal with the dynamic characteristics of different problems.

    \item \textbf{Algorithm structure innovation:} LLM is used to generate new algorithm structures or operators..

    \item \textbf{Decision support System:} Developing an LLM-based decision support tool to help decision-makers understand the obtained solutions, making the decision-making process more transparent and interpretable.
\end{itemize}

\section{Conclusion}

This paper discusses the application of LLM in multi-objective evolutionary optimization. By leveraging the power of a pre-trained LLM, we propose a novel way to use LLM as a black-box search operator in the MOEA framework. In addition, since LLM interactions consume a lot of time and resources, we employ hybrid strategies and flexible adjustment mechanisms to reduce interaction costs. Our experimental study demonstrates the effectiveness of our proposed method. Compared with the widely used MOEA, the performance of our framework is competitive, ranking first in some test instances.

\bibliography{MOEALLM}

\begin{thebibliography}{10}
\providecommand{\url}[1]{#1}
\csname url@samestyle\endcsname
\providecommand{\newblock}{\relax}
\providecommand{\bibinfo}[2]{#2}
\providecommand{\BIBentrySTDinterwordspacing}{\spaceskip=0pt\relax}
\providecommand{\BIBentryALTinterwordstretchfactor}{4}
\providecommand{\BIBentryALTinterwordspacing}{\spaceskip=\fontdimen2\font plus
\BIBentryALTinterwordstretchfactor\fontdimen3\font minus \fontdimen4\font\relax}
\providecommand{\BIBforeignlanguage}[2]{{%
\expandafter\ifx\csname l@#1\endcsname\relax
\typeout{** WARNING: IEEEtran.bst: No hyphenation pattern has been}%
\typeout{** loaded for the language `#1'. Using the pattern for}%
\typeout{** the default language instead.}%
\else
\language=\csname l@#1\endcsname
\fi
#2}}
\providecommand{\BIBdecl}{\relax}
\BIBdecl

\bibitem{introductionLLM1_1}
E.~Meyerson, M.~J. Nelson, H.~Bradley, A.~Moradi, A.~K. Hoover, and J.~Lehman, ``Language model crossover: Variation through few-shot prompting,'' \emph{arXiv preprint arXiv:2302.12170}, 2023.

\bibitem{introductionLLM1_2}
S.~Liu, C.~Chen, X.~Qu, K.~Tang, and Y.-S. Ong, ``Large language models as evolutionary optimizers,'' \emph{arXiv preprint arXiv:2310.19046}, 2023.

\bibitem{introductionLLM1_3}
F.~Liu, X.~Lin, Z.~Wang, S.~Yao, X.~Tong, M.~Yuan, and Q.~Zhang, ``Large language model for multi-objective evolutionary optimization,'' \emph{arXiv preprint arXiv:2310.12541}, 2023.

\bibitem{introductionLLM1_4}
C.~Yang, X.~Wang, Y.~Lu, H.~Liu, Q.~V. Le, D.~Zhou, and X.~Chen, ``Large language models as optimizers,'' \emph{arXiv preprint arXiv:2309.034091}, 2023.

\bibitem{introductionLLM1_5}
H.~Bradley, A.~Dai, H.~B. Teufel, J.~Zhang, K.~Oostermeijer, M.~Bellagente, J.~Clune, K.~Stanley, G.~Schott, and J.~Lehman, ``Qualitydiversity through ai feedback,'' \emph{in Proceedings of the 2nd Agent Learning in Open-Endedness Workshop and in 37th Annual Conference on Neural Information Processing Systems}, 2023.

\bibitem{relatedWorkMOEALLM1}
K.~Sanderson, ``Gpt-4 is here: what scientists think,'' \emph{Nature}, vol.~6, no.~2, pp. 182--197, 2023.

\bibitem{relatedWorkMOEALLM2}
J.~Chen, Z.~Liu, X.~Huang, C.~Wu, Q.~Liu, G.~Jiang, Y.~Pu, X.~C. Y.~Lei, and X.~W. et~al., ``When large language models meet personalization: Perspectives of challenges and opportunities,'' \emph{arXiv preprint arXiv:2307.16376}, 2023.

\bibitem{relatedWorkMOEALLM3}
P.-F. Guo, Y.-H. Chen, Y.-D. Tsai, and S.-D. Lin, ``Towards optimizing with large language models,'' \emph{arXiv preprint arXiv:2310.05204}, 2023.

\bibitem{relatedWorkMOEALLM9}
X.~Wu, Y.~Zhong, J.~Wu, and K.~C. Tan, ``As-llm: When algorithm selection meets large language model,'' \emph{arXiv preprint arXiv:2311.13184}, 2023.

\bibitem{relatedWorkMOEALLM10}
H.~Chen, G.~E. Constante-Flores, and C.~Li, ``Diagnosing infeasible optimization problems using large language models,'' \emph{arXiv preprint arXiv:2308.12923}, 2023.

\bibitem{relatedWorkMOEALLM11}
M.~Pluhacek, A.~Kazikova, T.~Kadavy, A.~Viktorin, and R.~Senkerik, ``Leveraging large language models for the generation of novel metaheuristic optimization algorithms,'' \emph{in Proceedings of the Companion Conference on Genetic and Evolutionary Computation}, pp. 1812--1820, 2023.

\bibitem{relatedWorkMOEALLM12}
H.~Bradley, H.~Fan, T.~Galanos, R.~Zhou, D.~Scott, and J.~Lehman, ``The openelm library: Leveraging progress in language models for novel evolutionary algorithms.''

\bibitem{relatedWorkLLM3}
Yang, C., Wang, X., Lu, Y., Liu, H., Le, Q.~V., Zhou, and D, ``Large language models as optimizers,'' \emph{arXiv preprint arXiv:2309.03409}, 2023.

\bibitem{relatedWorkLLM4}
Liu, S., Chen, C., Qu, X., Tang, K., Ong, and Y.~S, ``Large language models as evolutionary optimizers,'' \emph{arXiv preprint arXiv:2310.19046}, 2023.

\bibitem{relatedWorkMOEALLM6}
F.~Liu, X.~Lin, Z.~Wang, S.~Yao, X.~Tong, M.~Yuan, and Q.~Zhang, ``Large language model for multi-objective evolutionary optimization,'' \emph{arXiv preprint arXiv:2310.12541}, 2023.

\bibitem{relatedWorkMOEALLM7}
J.~Lehman and K.~O. Stanley, ``Evolving a diversity of virtual creatures through novelty search and local competition,'' \emph{in Proceedings of the 13th Annual Conference on Genetic and Evolutionary Computation}, pp. 211--218, 2011.

\bibitem{relatedWorkMOEALLM8}
H.~Bradley, A.~Dai, H.~B. Teufel, J.~Zhang, K.~Oostermeijer, M.~Bellagente, J.~Clune, K.~Stanley, G.~Schott, and J.~Lehman, ``Qualitydiversity through ai feedback,'' \emph{in Proceedings of the 2nd Agent Learning in Open-Endedness Workshop and in 37th Annual Conference on Neural Information Processing Systems}, 2023.

\bibitem{relatedWorkLLM2}
Guo, P.~F., Chen, Y.~H., Tsai, Y.~D., Lin, and S.~D, ``Towards optimizing with large language models,'' \emph{arXiv preprint arXiv:2310.05204}, 2023.

\bibitem{miettinen2012nonlinear}
Miettinen and Kaisa, \emph{Nonlinear multiobjective optimization}.\hskip 1em plus 0.5em minus 0.4em\relax Springer Science \& Business Media, 2012, vol.~12.

\bibitem{Nsgaii}
Srinivas, N., Deb, and Kalyanmoy, ``Muiltiobjective optimization using nondominated sorting in genetic algorithms,'' \emph{Evolutionary Computation}, vol.~2, no.~3, pp. 221--248, 1994.

\bibitem{deb1995multi}
Deb, Kalyanmoy, Srinivasan, and Aravind, ``Multi-objective genetic algorithms: Problem difficulties and construction of test problems,'' in \emph{Proceedings of the International Conference on Genetic Algorithms}, 1995, pp. 109--115.

\bibitem{xiong2023can}
Xiong, Miao, Hu, Zhiyuan, Lu, Xinyang, Li, Yifei, Fu, Jie, He, Junxian, Hooi, and Bryan, ``Can llms express their uncertainty? an empirical evaluation of confidence elicitation in llms,'' \emph{arXiv preprint arXiv:2306.13063}, 2023.

\bibitem{li2008multiobjective}
Li, Hui, Zhang, and Qingfu, ``Multiobjective optimization problems with complicated pareto sets and \text{MOEA/D} and \text{NSGA-II},'' \emph{IEEE transactions on evolutionary computation}, vol.~13, no.~2, pp. 284--302, 2008.

\bibitem{zitzler2000comparison}
Zitzler, Eckart, Deb, Kalyanmoy, Thiele, and Lothar, ``Comparison of multiobjective evolutionary algorithms: Empirical results,'' \emph{Evolutionary computation}, vol.~8, no.~2, pp. 173--195, 2000.

\bibitem{Arsbx}
Pan, Linqiang, Xu, Wenting, Li, Lianghao, He, Cheng, Cheng, and Ran, ``Adaptive simulated binary crossover for rotated multi-objective optimization,'' \emph{Swarm and Evolutionary Computation}, vol.~60, p. 100759, 08 2020.

\bibitem{zhang2007moea}
Zhang, Qingfu, Li, and Hui, ``Moea/d: A multiobjective evolutionary algorithm based on decomposition,'' \emph{IEEE Transactions on Evolutionary Computation}, vol.~11, no.~6, pp. 712--731, 2007.

\bibitem{deb2014nsgaiii}
Deb, Kalyanmoy, Jain, and Himanshu, ``An evolutionary many-objective optimization algorithm using reference-point-based nondominated sorting approach and part i: solving problems with box constraints,'' \emph{IEEE Transactions on Evolutionary Computation}, vol.~18, no.~4, pp. 577--601, 2014.

\bibitem{zhang2009moea}
Zhang, Qingfu, Zhou, Aimin, Zhao, Shengxiang, Suganthan, P.~Nagaratnam, Liu, Wenyin, Zhang, and Weiyin, ``Multiobjective optimization test instances for the cec 2009 special session and competition,'' in \emph{IEEE Congress on Evolutionary Computation}.\hskip 1em plus 0.5em minus 0.4em\relax IEEE, 2009, pp. 1--30.

\bibitem{zhang2020moead_dqn}
Zhang, Hui, Zhou, Aimin, Zhang, and Qingfu, ``A deep reinforcement learning based multiobjective evolutionary algorithm using decomposition,'' \emph{IEEE Transactions on Evolutionary Computation}, vol.~24, no.~3, pp. 494--507, 2020.

\bibitem{tian2017platemo}
Tian, Ye, Cheng, Ran, Zhang, Xingyi, Jin, and Yaochu, ``Platemo: A matlab platform for evolutionary multi-objective optimization [educational forum],'' \emph{IEEE Computational Intelligence Magazine}, vol.~12, no.~4, pp. 73--87, 2017.

\end{thebibliography}
\bibliographystyle{IEEEtran}

\end{document}